%% file: COLI_template.tex
\documentclass[final]{clv2025}

\jvol{vv}
\jnum{nn}
\jyear{2025}

\usepackage{amsmath}
\usepackage{booktabs}
\usepackage{algpseudocode, algorithm}
\usepackage{amsmath, amssymb, amsthm, hyperref}
\usepackage{mathtools} 
\usepackage{multirow}
\usepackage{pifont}
\usepackage{booktabs}
\usepackage{adjustbox}
\usepackage{colortbl}
\usepackage{makecell}
\usepackage{subcaption}
\usepackage[most]{tcolorbox}
\usepackage{xcolor}
\usepackage{wrapfig}
\usepackage{pdflscape}
\usepackage{graphicx}
\usepackage{rotating}
\usepackage{afterpage}
\usepackage[T1]{fontenc}

\usepackage[utf8]{inputenc}

\usepackage{microtype}

\usepackage{inconsolata}
\usepackage{enumitem}
\usepackage{graphicx}

\definecolor{LightGray}{HTML}{C0C0C0}
\definecolor{OliveGreen}{rgb}{0.0, 0.5, 0.0}
\usepackage{hyperref}
\usepackage{cleveref}

\crefformat{section}{\S#2#1#3}
\crefformat{subsection}{\S#2#1#3}
\crefformat{subsubsection}{\S#2#1#3}
\crefrangeformat{section}{\S\S#3#1#4 to~#5#2#6}
\crefmultiformat{section}{\S\S#2#1#3}{ and~#2#1#3}{, #2#1#3}{ and~#2#1#3}

\newcommand{\method}{\texttt{HIDE}}

\newcommand{\cmark}{\ding{51}} 
\newcommand{\xmark}{\ding{55}} 

\runningtitle{Hallucination Detection via Decoupled
Representations}
\runningauthor{Chatterjee, Goel, Chakraborty}

\makeatletter
\renewcommand*{\@fnsymbol}[1]{%
  \ensuremath{%
    \ifcase#1\or               
      \ast       
    \or\dagger   
    \or\ddagger  
    \or\S        
    \or\P        
    \or\dagger\dagger 
    \or\S\S      
    \else\@arabic#1%
  \fi}}
\makeatother

\begin{document}

\title{\text{\method~and Seek}: Detecting Hallucinations in Language Models via Decoupled Representations}

\author{Anwoy Chatterjee\thanks{Equal contribution}\thanks{Corresponding author: \email{anwoy.chatterjee@ee.iitd.ac.in}}, Yash Goel\footnotemark[1], Tanmoy Chakraborty}

\affilblock{
    \affil{Indian Institute of Technology Delhi, India}
}

\maketitle

\input{files/abstract}

\input{files/intro}
\input{files/related_work}
\input{files/preliminary}
\input{files/method}

\input{files/results_table}

\input{files/exp_setup}
\input{files/results}

\input{files/ablations}
\input{files/error_analysis}

\input{files/conclusion}

\appendix
\input{files/proofs}
\input{files/exact_match_results}

\input{files/Prompting_techniques}
\newpage
\bibliographystyle{compling}
\bibliography{COLI_template}

\end{document}

%% file: files/abstract.tex
\begin{abstract}
Contemporary Language Models (LMs), while impressively fluent, often generate content that is factually incorrect or unfaithful to the input context --- a critical issue commonly referred to as `hallucination'. This tendency of LMs to generate hallucinated content undermines their reliability, especially because these fabrications are often highly convincing and therefore difficult to detect.
While several existing methods attempt to detect hallucinations, most rely on analyzing multiple generations per input, leading to increased computational cost and latency. To address this, we propose a single-pass, training-free approach for effective \textbf{H}allucination detect\textbf{I}on via \textbf{D}ecoupled r\textbf{E}presentations (\textit{\texttt{\textbf{HIDE}}}). 
Our approach leverages the hypothesis that hallucinations result from a statistical decoupling between an LM's internal representations of input context and its generated output. We quantify this decoupling using the Hilbert-Schmidt Independence Criterion (HSIC) applied to hidden-state representations extracted while generating the output sequence.
We conduct extensive experiments on four diverse question answering datasets, evaluating both faithfulness and factuality hallucinations across six open-source LMs of varying scales and properties. Our results demonstrate that \method~outperforms other single-pass methods in almost all settings, achieving an average relative improvement of $\sim 29\%$ in AUC-ROC over the best-performing single-pass strategy across various models and datasets. Additionally, \method~shows competitive and often superior performance with multi-pass state-of-the-art methods, obtaining an average relative improvement of $\sim 3\%$ in AUC-ROC while consuming $\sim 51\%$ less computation time. Our findings highlight the effectiveness of exploiting internal representation decoupling in LMs for efficient and practical hallucination detection.\footnote{The code to reproduce our results is open-sourced: \url{https://github.com/C-anwoy/HIDE}.}

\end{abstract}


%% file: files/intro.tex
\section{Introduction}

Our evolving relationship with \textit{truth} in the age of artificial intelligence is perhaps most seriously challenged by modern Language Models (LMs) \cite{Brown2020, Chowdhery2022, Touvron2023}. These models, capable of generating text with remarkable human-like fluency, come with a troubling limitation: \textit{they can be confidently and convincingly wrong}. This phenomenon, often known as \textit{hallucination} \cite{ji_survey} -- where models produce factually incorrect, nonsensical or misleading content that appears fluent and convincing -- poses a serious challenge to their reliable and trustworthy use \cite{zhang2023sirenssongaiocean, 10.1145/3531146.3533088, Huang_2025, maynez-etal-2020-faithfulness}. 
With these systems now being integrated across various platforms and used en masse, the chances increase that users may take hallucinated content at face value, often without realizing it. As a result, the line between genuine knowledge and convincing fabrication becomes dangerously easy to blur. 

Hallucinations come in different forms. \citet{Huang_2025} distinguished between \textit{faithfulness hallucinations}, where the output directly contradicts the provided source, and \textit{factuality hallucinations}, where the the model-generated output diverges from real-world facts. The origins of this issue are often traced to training data, model architecture, and inference methods \cite{lee_deduplicating_2022, onoe_entity_2022, zheng_why_2023}. 
While the term `hallucination' itself is often debated for its anthropomorphic associations \citep{halu_debate}, the underlying issue of unreliability poses significant risks, especially in critical domains like healthcare and finance. Addressing this by developing robust detection methods is therefore crucial for the trustworthy deployment of LMs.

\begin{figure*}[t!]
    \centering
    \includegraphics[width=\linewidth]{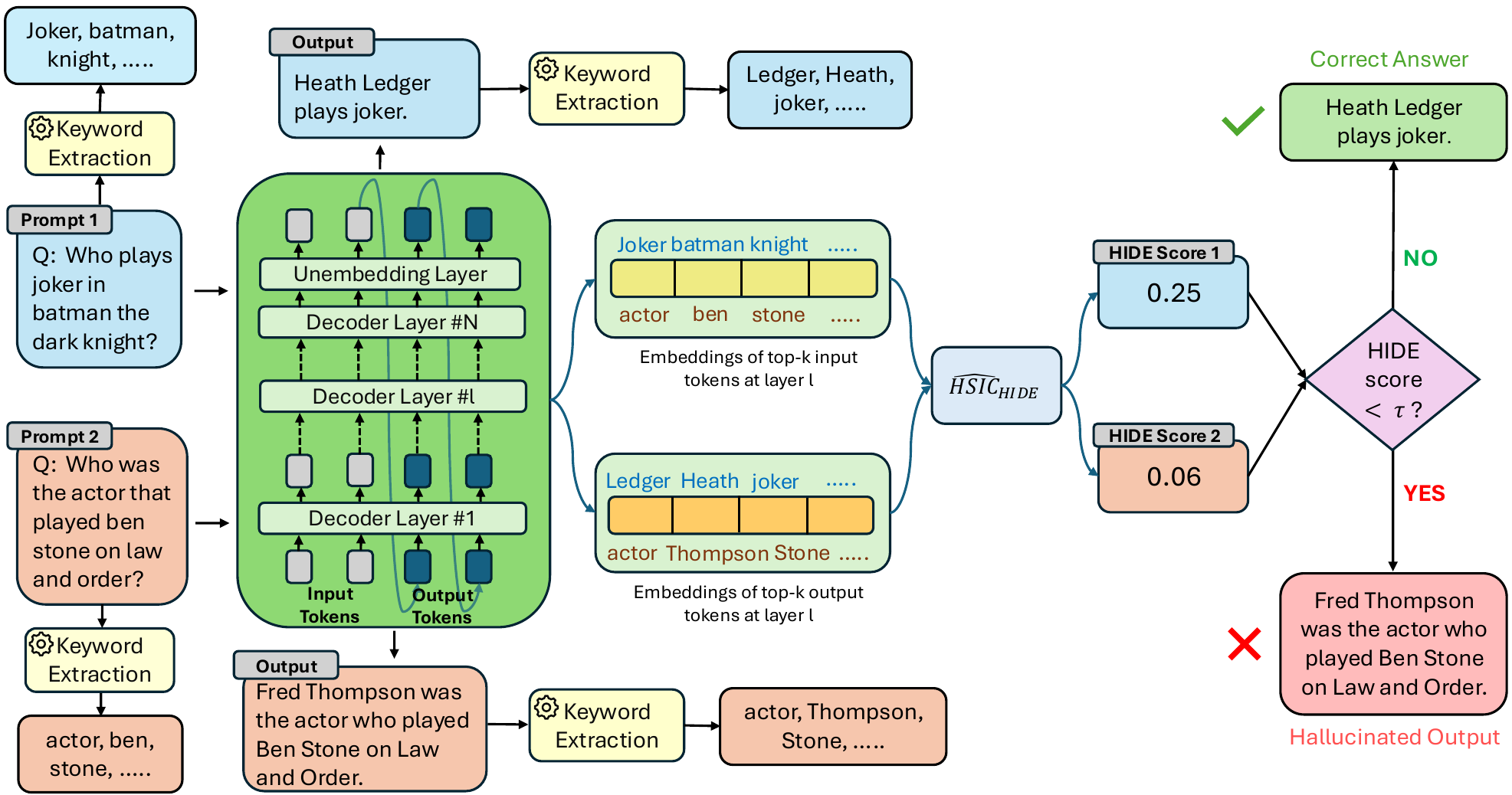}
    \caption{Our proposed method, \method~is a single-pass, training-free approach for hallucination detection. For a given input prompt, \method~extracts representative tokens from both the input and the LM-generated output. It then computes a \method~score using an adapted variant of unbiased HSIC on the hidden state embeddings of these token sets from a specific LM layer. This score, reflecting input-output representational dependence, is compared against a threshold to detect hallucinated generation.
    }
    \label{fig:intro}
\end{figure*}

A common approach to detecting hallucination is to measure the model's uncertainty using metrics like perplexity \citep{ren_out--distribution_2023}, entropy \citep{malinin_uncertainty_2021}, or energy-based scores \citep{liu_energy-based_2020}, based on the assumption that lower confidence signals a higher likelihood of hallucination. While sometimes useful, uncertainty-based methods often perform poorly in detecting hallucinations, as LMs show the tendency to remain highly confident even when hallucinating \citep{simhi2025trustmeimwrong}. Another approach is to measure consistency by generating multiple answers to the same prompt and comparing them using lexical similarity \citep{lin_towards_2022}, semantic similarity (e.g., BERTScore, NLI), QA-based agreement, $n$-gram overlap, or LLM-based assessment \citep{manakul2023selfcheckgptzeroresourceblackboxhallucination}. \citet{chen_inside:_2024} also introduced a white-box method that quantifies output diversity in the latent space of LMs' internal representations \footnote{We have used the terms \textit{internal representation} and \textit{hidden state} interchangeably in this paper.}. Despite being effective, these methods require generating multiple outputs for every input, making them computationally expensive and challenging for real-time detection. With concurrent advancements in our understanding of how different mechanisms related to hallucinations are facilitated within LMs, various works \citep{azaria2023internalstatellmknows, ji-etal-2024-llm, duan2024llmsknowhallucinationempirical} have shown that the hidden states of LMs hold certain cues when it is hallucinating. While the aim of these works is to investigate the hallucination phenomenon, and some of them separately train probing classifiers on top of the hidden states, we aim to use these representations for hallucination detection in a training-free setting during inference. In this work, we, therefore, attempt to address the issue of higher latency in detecting hallucinations during inference, and investigate the research question: \textit{Can we leverage the internal representations of LMs to detect hallucinations effectively and efficiently in a single-pass \footnote{In this paper, by \textit{single-pass} we refer to requiring no additional model runs beyond the standard auto-regressive generation of a single output sequence for a given input.}, training-free setting?} 

To this end, we propose \textbf{\method} (\textbf{H}allucination Detect\textbf{I}on via \textbf{D}ecoupled R\textbf{E}presentations), a single-pass method that quantifies hallucination by measuring the statistical dependence between an LM's internal representations of input and output sequences. Figure~\ref{fig:intro} illustrates how \method~extracts hidden states corresponding to key tokens from both input and output, then applies an adapted variant of the Hilbert-Schmidt Independence Criterion (HSIC) to these representations to derive a score indicative of representational coupling. A low score suggests decoupling, signaling a potential hallucination. Unlike multi-pass approaches, \method~only requires a single generation for each input, making it suitable for deployment in real-world systems where efficiency and latency are key considerations.

We conduct extensive experiments across six open-source LMs of varying scales and properties, including base and instruction-tuned variants of Llama-3.2-3B, Llama-3-8B~\citep{grattafiori2024llama3herdmodels}, and Gemma-2-9B~\citep{gemmateam2024gemma2improvingopen}, on diverse question answering datasets, like SQuAD \citep{rajpurkar2016squad100000questionsmachine} and RACE \citep{lai-etal-2017-race} to assess \method's ability to detect faithfulness hallucinations, and Natural Questions \citep{kwiatkowski-etal-2019-natural} and TriviaQA \citep{joshi2017triviaqalargescaledistantly} to test the effectiveness of \method\ for factuality hallucination. We observe that:
\begin{itemize}[nosep, wide, labelwidth=!, labelindent=1.5pt]
    \item \method~effectively captures the representational decoupling, hypothesized to underlie both faithfulness and factuality hallucinations, demonstrating its versatility across hallucination categories (\cref{sec:faithful_results,sec:factual_results}).
    \item \method~consistently outperforms other single-pass detection techniques, utilizing uncertainty-detection measures like perplexity and energy-based scores, in distinguishing hallucinated content in almost all settings. It achieves an average relative improvement of $\sim 29\%$ in AUC-ROC over the best performing single-pass method across various considered models and datasets (\cref{sec:results}).
    \item Despite its single-pass nature, \method~achieves performance that is competitive with, and in several settings superior to, the state-of-the-art multi-pass methods like Eigenscore. Moreover, \method~shows an average improvement of $\sim 3\%$ in AUC-ROC over the best multi-pass detection method across different settings, while reducing the average computation time by $\sim 51\%$ (\cref{sec:compute_time}).
\end{itemize}

Our analyses also reveal that the performance of \method~is almost agnostic to layer and kernel selection, 
further emphasizing \method's robustness and practical applicability (\cref{sec:ablations}). Our findings suggest that measuring internal representational dependence is a promising direction for developing efficient and reliable hallucination detection systems, aiding the research community in building more trustworthy LMs.

%% file: files/related_work.tex
\section{Background and Related Work}
\label{sec:related_work}

Though hallucination is often deemed to be essential for the creativity of LMs \citep{math11102320, 10.1145/3652102}, it remains one of the most frequently criticized shortcomings of these models. While some studies argue that hallucinations are an inherent consequence of their stochastic nature, often known as the `stochastic parrots' phenomenon \citep{10.1145/3442188.3445922}, many others attribute the issue primarily to artifacts in the data that these models ingest. There is a substantial body of research addressing diverse aspects of hallucination in LMs -- we briefly summarize a few of these works related to two major dimensions: sources of hallucination and hallucination detection.

\subsection{Sources of Hallucination} 

Hallucinations in LMs have been attributed to a complex interplay of factors spanning data quality, model architecture, and inference processes. On the data front, pre-training on vast and often imperfect corpora introduces risks of propagating misinformation, factual inaccuracies, and social biases~\citep{lin-etal-2022-truthfulqa, lee_deduplicating_2022, hernandez2022scalinglawsinterpretabilitylearning, data_paullada_2021, narayanan-venkit-etal-2023-nationality, ladhak-etal-2023-pre, kasai2024realtimeqawhatsanswer}. The ever-increasing scale of training data, while enabling greater generalization, can also amplify errors and result in LMs memorizing, rather than reasoning about information.

At the level of model training and design, certain architectural limitations -- such as insufficient bidirectional representation or attention mechanism failures -- can introduce systematic vulnerabilities to hallucination~\citep{li2023batgptbidirectionalautoregessivetalker, liu2023exposingattentionglitchesflipflop, wang_exposure_2020, sharma_towards_2023}. Moreover, the alignment process that tunes LMs to follow human preferences may occasionally introduce a mismatch between model capabilities and desired behaviors, leading to unexpected model outputs.

Inference strategies, especially those introducing stochasticity or prioritizing diversity (such as top-$k$ or nucleus sampling), have also been linked to hallucination~\citep{fan-etal-2018-hierarchical, holtzman_curious_2019, dziri-etal-2021-neural, chuang2024doladecodingcontrastinglayers}. Additional challenges arise from limitations in context integration and representational bottlenecks in output layers~\citep{liu_instruction_2023, shi2023trustingevidencehallucinatecontextaware, yang2018breakingsoftmaxbottleneckhighrank, chang_softmax_2022}. Together, these findings indicate that hallucinations arise not from a single source but from interactions among data, architecture, and inference.

\subsection{Hallucination Detection} 

Given the broad impact of hallucination, a wide range of detection strategies has been proposed. Many early efforts focused on evaluating model confidence, with the intuition that LMs are less certain when hallucinating. Metrics such as perplexity, entropy, and energy-based scores~\citep{ren_out--distribution_2023, malinin_uncertainty_2021, liu_energy-based_2020} have been explored for this purpose. Another family of approaches leverages output consistency: by sampling multiple generations for the same prompt, one can measure lexical or semantic agreement across responses~\citep{lin-etal-2022-truthfulqa, durmus-etal-2020-feqa, scialom-etal-2021-questeval}. High variability is often taken as a proxy for hallucination, and techniques such as BERTScore, NLI-based classification, and question-answering-based evaluation have all been applied~\citep{falke-etal-2019-ranking, maynez-etal-2020-faithfulness, nan-etal-2021-entity, chen2024complexclaimverificationevidence}. However, these multi-pass strategies are computationally demanding and poorly suited to low-latency settings, as they require generating and analyzing numerous outputs for each input.

Other detection frameworks rely on verifying factuality by grounding model outputs in external knowledge sources~\citep{min2023factscorefinegrainedatomicevaluation, huo2023retrievingsupportingevidencellms, chern2023factoolfactualitydetectiongenerative, chen2024complexclaimverificationevidence}. While these methods can be effective, they require access to curated external databases and ongoing maintenance, and may not generalize across domains.

More recently, studies have explored the use of LMs themselves as evaluators -- either via prompt engineering that elicits self-critique, or by prompting additional LMs to judge the factuality or faithfulness of generated content~\citep{xiong2024llmsexpressuncertaintyempirical, kadavath2022languagemodelsmostlyknow, manakul2023selfcheckgptzeroresourceblackboxhallucination, chiang2023largelanguagemodelsalternative, luo2023chatgptfactualinconsistencyevaluator, laban2023llmsfactualreasonersinsights, miao2023selfcheckusingllmszeroshot}. These meta-evaluation techniques provide flexibility in black-box settings, but can inherit the blind spots of the underlying models and are themselves subject to hallucination.

A comparatively smaller but growing line of work examines the internal dynamics of language models, investigating whether the hidden states or intermediate representations carry signals that are indicative of hallucination~\citep{chen2024complexclaimverificationevidence, ji-etal-2024-llm}. These studies have shown that certain patterns of representational diversity of the hidden states may be predictive of hallucinated content. However, most prior work in this area has either required separate classifier training on internal states or operated in multi-pass setups, limiting their practicality for inference-time detection.

\subsection{Background of Hallucination Detection Measures} \label{sec:background}

Here we briefly discuss the mathematical formulation of a few hallucination detection measures, which are used as baselines in this work. Since our proposed method doesn't involve any training, we compare only against other training-free methods.

Assume that $\mathbf{S}_{in} = \{s_1, s_2, \ldots, s_I\}$ is the input prompt, and $\mathbf{S}_{out} = \{t_1, t_2, \ldots, t_O\}$ is the corresponding output generated by the LM, parameterized by $\theta$, where $\mathbf{S}_{out}$ is a sequence of $O$ tokens. In case of measures requiring multiple generations for the same input $\mathbf{S}_{in} $, assume $\mathcal{S} =[\mathbf{S}_{out}^1, \mathbf{S}_{out}^2, ..., \mathbf{S}_{out}^N]$ to be the collection of $N$ generations, where diversity is ensured through stochastic sampling strategies, like top-$k$, nucleus or temperature sampling. Also, $\mathbb{P}_{LM}(.)$ denotes the probability distribution generated by the LM for any position in the sequence. 
\paragraph{\textbf{Perplexity}}
 Perplexity, as a token-level uncertainty measure, can be utilized for evaluating potential hallucinations \citep{ren_out--distribution_2023}. Perplexity is monotonically related to the mean negative log-likelihood (MNLL) of the output sequence, and hence MNLL can be utilized as a tractable proxy for perplexity as a hallucination detection measure \citep{chen_inside:_2024}. Formally, MNLL can be defined as: $\text{MNLL}(\mathbf{S}_{out}|\mathbf{S}_{in}; \theta) = -\frac{1}{O}\sum_{i=1} ^{O}\log \mathbb{P}_{LM}(t_i|t_{<i}, \mathbf{S}_{in})$. A lower value of this measure indicates that the model is more confident about its generation. 

\paragraph{\textbf{Energy Score}}
The Energy Score \citep{liu_energy-based_2020} approaches hallucination detection from a different perspective by leveraging energy-based models. Unlike softmax-based confidence scores, energy scores are aligned with the probability density of the inputs. 
Let $\mathbb{L}_{LM}(.)$ denote the logit distribution over the set of tokens in the vocabulary $\mathcal{V}$, generated by the LM, such that $\mathbb{P}_{LM}(.) = \text{softmax}(\mathbb{L}_{LM}(.))$. The energy score, for a temperature parameter $T$, can then be defined as: $\text{Energy}(\mathbf{S}_{in}; \theta) = -T \cdot \log\sum_{t \in \mathcal{V}} e^{\mathbb{L}_{LM}(t|\mathbf{S}_{in})/T}$. The energy score is generally used for out-of-distribution detection, where samples with higher energies can be interpreted as datapoints with a lower likelihood of occurrence.

\paragraph{\textbf{Length-Normalized Entropy}} Unlike perplexity which utilizes a single generation for quantifying sequence-level uncertainty, length-normalized entropy (LN-Entropy) uses multiple generations to measure the degree of hallucination~\citep{malinin_uncertainty_2021}. LN-Entropy is defined as: $\text{LNE}(\mathcal{S}|\mathbf{S}_{in}; \theta) = -\mathbb{E}_{\mathbf{S}_{out}\in \mathcal{S}}\frac{1}{|\mathbf{S}_{out}|}\sum_{i=1}^{|\mathbf{S}_{out}|}\log \mathbb{P}_{LM}(t_i|t_{<i}, \mathbf{S}_{in})$. It is assumed that when an LM generates hallucinated content, it is typically uncertain, resulting in an output distribution with higher entropy \cite{kadavath2022languagemodelsmostlyknow}.

\paragraph{\textbf{Lexical Similarity}} Lexical Similarity (LS) \cite{lin_towards_2022} aims to quantify the similarities between the multiple generations from an LM corresponding to the same input prompt, with the assumption that LS will be low in case of hallucinations. It can formally defined as: 
$\text{LS}(\mathcal{S}|\mathbf{S}_{in}; \theta) = \frac{2}{N(N-1)}\sum_{i=1}^{N}\sum_{j=i+1}^{N}sim(\mathbf{S}_{out}^i, \mathbf{S}_{out}^j)$, where, $sim(\cdot, \cdot)$ can be any function to quantify similarity, like Rouge-L \cite{lin-2004-rouge}.

\paragraph{\textbf{EigenScore}}
EigenScore \citep{chen_inside:_2024} leverages the hidden states of LMs for hallucination detection. It measures semantic diversity in the latent space of LMs' internal representations by calculating the log-determinant of the covariance matrix of multiple output representations: $\text{ES}(\mathcal{S}|\mathbf{S}_{in}; \theta)  = \frac{1}{N}\log\det(\Sigma + \alpha \cdot \mathbb{I}_N) = \frac{1}{N}\sum_{i=1}^{N}\log(\lambda_i)$, where, the covariance matrix of the output representations is given by: $\Sigma = Z^T \cdot C_d \cdot Z$ with $Z = [z_1, z_2, \cdots, z_N] \in \mathbb{R}^{d\times N}$ being the embedding matrix of the $N$ generated outputs and $C_d = \mathbb{I}_d - \frac{1}{d}\mathbf{1}_d\mathbf{1}_d^T$ being the centering matrix. The set of eigenvalues of the regularized covariance matrix $\Sigma + \alpha \cdot \mathbb{I}$ is given by $\lambda = \{\lambda_1, \lambda_2, \cdots, \lambda_N\}$. It is assumed that when the LM is not hallucinating, the representations of the outputs will be highly correlated, thereby having $\lambda_i=0$ for most $i$'s.

Table~\ref{tab:evaluation_methods} presents a summary of the characteristics of the discussed hallucination detection measures, and compares our proposed method with these techniques.

\begin{table}[t!]
\centering
\resizebox{0.8\linewidth}{!}{
\begin{tabular}{lccc}
\toprule
\textbf{Method} & \makecell{\textbf{Uses} \\ \textbf{Uncertainty}} & \makecell{\textbf{Checks} \\ \textbf{Consistency}} & \makecell{\textbf{No. of Generations} \\ \textbf{Required}} \\
\midrule
Perplexity       & \cmark & \xmark & 1 \\
Energy        & \cmark & \xmark & 1 \\
Length-Normalized Entropy     & \cmark & \xmark & $N$ \\
Lexical Similarity & \xmark & \cmark & $N$ \\
Eigenscore & \xmark & \cmark & $N$ \\
\rowcolor{gray!10}
\textbf{\method~(Ours)}        & \xmark & \xmark & \text{1} \\
\bottomrule
\end{tabular}
}
\caption{Comparison of some popular training-free hallucination detection techniques. Most methods rely on either model uncertainty or consistency across multiple generations ($N$) -- approaches relying on the latter show increased latency and compute cost. \method, on the other hand, requires only a single generation and leverages internal representation independence, offering a lightweight yet effective alternative.}

\label{tab:evaluation_methods}
\end{table}

%% file: files/preliminary.tex
\section{Theoretical Foundations}
\label{sec:foundations_hsic}
We propose that linguistic hallucination in LMs often stems from a statistical decoupling between the model's internal representation of the input and that of the generated output. Our experiments show that, while sound generation preserves a robust statistical dependency between these representations, hallucination is characterized by a measurable weakening of this dependence, as quantified by our method (\cref{sec:irid,sec:results}). To quantify this phenomenon, we leverage the Hilbert-Schmidt Independence Criterion (HSIC), a non-parametric measure that can capture complex, non-linear relationships in high-dimensional spaces, such as LMs' hidden states. We first introduce HSIC and discuss its properties, motivating its adaptation to our method for hallucination detection. 

\subsection{Measuring Statistical Dependence using HSIC}
\label{subsec:hsic_rkhs}
Let's assume that we want to assess the statistical dependence between two random variables $X$ and $Y$, taking values in separable topological spaces $\mathcal{X}$ and $\mathcal{Y}$, respectively. The statistical relationship between $X$ and $Y$ is governed by their joint probability distribution $P_{XY}$. HSIC quantifies this dependence by comparing $P_{XY}$ to the product of the marginals, $P_X P_Y$, within Reproducing Kernel Hilbert Spaces (RKHSs). An RKHS $\mathcal{F}_k$, associated with a kernel $k: \mathcal{X} \times \mathcal{X} \to \mathbb{R}$ allows implicit mapping to a feature space via $\phi: \mathcal{X} \to \mathcal{F}_k$, where $k(x, x') = \langle \phi(x), \phi(x') \rangle_{\mathcal{F}_k}$, with $\langle \cdot, \cdot \rangle_{\mathcal{F}_k}$ denoting the inner product in the RKHS $\mathcal{F}_k$. This mapping facilitates the detection of non-linear dependencies.

Given two RKHSs, $\mathcal{F}$ with kernel $k_X$ for variable $X$, and $\mathcal{G}$ with kernel $k_Y$ for variable $Y$ (corresponding to feature maps $\phi$ and $\psi$ respectively), the cross-covariance operator $C_{XY}: \mathcal{G} \to \mathcal{F}$ is defined as:
\begin{equation*}\small
\begin{small}
C_{XY} \coloneqq \mathbb{E}_{XY}[\phi(X) \otimes \psi(Y)] - \mathbb{E}_X[\phi(X)] \otimes \mathbb{E}_Y[\psi(Y)]
\end{small}
\end{equation*}
where $\otimes$ denotes the tensor product. Intuitively, $C_{XY}$ captures the covariance between the representations of $X$ and $Y$ in their respective RKHSs. If $X$ and $Y$ are independent, $C_{XY}$ is the zero operator. 

The Hilbert-Schmidt norm of an operator $C: \mathcal{G} \to \mathcal{F}$, denoted by $\|C\|_{HS}$, is a generalization of the Frobenius norm for matrices and can be defined as $\|C\|_{HS}^2 = \sum_{i,j} |\langle C g_j, f_i \rangle_{\mathcal{F}}|^2$, where $\{f_i\}$ and $\{g_j\}$ are orthonormal bases of $\mathcal{F}$ and $\mathcal{G}$, respectively. It essentially measures the \textit{magnitude} of the operator.

\begin{definition}
\label{def:hsic}
\textit{HSIC is the squared Hilbert-Schmidt norm of the cross-covariance operator $C_{XY}$:}
\begin{equation}
\text{\textit{HSIC}}(X,Y; k_X, k_Y) \coloneqq \|C_{XY}\|^2_{HS}
\end{equation}
\end{definition}

\noindent A crucial property of HSIC is its relationship with statistical independence, formalized as follows:

\begin{lemma}[\citet{gretton05a}]
\label{lemma:hsic_independence_merged}
\textit{Let $k_X$ and $k_Y$ be characteristic kernels on $\mathcal{X}$ and $\mathcal{Y}$, respectively. Then $\operatorname{HSIC}(X,Y; k_X, k_Y) = 0$ if and only if $X$ and $Y$ are statistically independent.}
\end{lemma}
A kernel $k$ is termed characteristic if the mapping from probability measures on its domain to mean elements in its RKHS, $\mu \mapsto \mathbb{E}_{X \sim \mu}[\phi(X)]$, is injective \citep{JMLR:v11:sriperumbudur10a}. This ensures that the mean embedding $\mathbb{E}[\phi(X)]$ uniquely determines $P_X$. For our method, we use the Radial Basis Function (RBF) kernel, $k_{\text{RBF}}(x, x') = \exp(-\gamma \|x - x'\|_2^2)$ for a bandwidth parameter $\gamma > 0$.

\begin{lemma}[\citet{5122}]
\label{lemma:rbf_characteristic_merged}
\textit{The RBF kernel is characteristic on $\mathbb{R}^d$.}
\end{lemma}
The conjunction of Lemma~\ref{lemma:hsic_independence_merged} and Lemma~\ref{lemma:rbf_characteristic_merged} establishes HSIC, when computed with RBF kernels, as a robust and theoretically sound metric for detecting statistical independence, forming the cornerstone of our approach.

\subsection{Empirical Estimation of HSIC}
\label{subsec:empirical_hsic_foundation}
In practice, we rely on empirical estimates of HSIC to quantify statistical independence from a finite set of $n$ samples. Given samples $\{x_1, \dots, x_n\}$ from the distribution of $X$ and $\{y_1, \dots, y_n\}$ from the distribution of $Y$, we form $n \times n$ Gram matrices $\mathbf{K}_X$ and $\mathbf{K}_Y$, where $\mathbf{K}_{X,ij} = k_X(x_i, x_j)$ and $\mathbf{K}_{Y,ij} = k_Y(y_i, y_j)$.

A common, though biased, empirical estimator for HSIC is \citep{gretton05a}:

\begin{equation}
\widehat{\operatorname{HSIC}}_b(X,Y) = \frac{1}{(n-1)^2} \text{Tr}(\mathbf{K}_X \mathbf{H} \mathbf{K}_Y \mathbf{H})
\label{eq:biased_HSIC_foundation}
\end{equation}
where $\mathbf{H} = \mathbf{I} - \frac{1}{n}\mathbf{1}\mathbf{1}^T$ is the $n \times n$ centering matrix, $\mathbf{I}$ is the identity matrix, and $\mathbf{1}$ is an $n \times 1$ column vector of ones.

For improved performance, particularly with smaller $n$, an unbiased estimator is preferred \citep{JMLR:v13:song12a}. The V-statistic based unbiased estimator is given by:
\begin{multline}
\widehat{\operatorname{HSIC}}_u(X,Y) = \frac{1}{n(n-3)} \bigg[ \text{Tr}(\tilde{\mathbf{K}}_X\tilde{\mathbf{K}}_Y) +   \frac{(\mathbf{1}^T \tilde{\mathbf{K}}_X \mathbf{1}) (\mathbf{1}^T \tilde{\mathbf{K}}_Y \mathbf{1})}{(n-1)(n-2)} -  \frac{2}{n-2} \mathbf{1}^T \tilde{\mathbf{K}}_X\tilde{\mathbf{K}}_Y \mathbf{1} \bigg]
\label{eq:unbiased_HSIC_foundation}
\end{multline}
Here, $n$ is the number of samples. The matrices $\tilde{\mathbf{K}}_X$ and $\tilde{\mathbf{K}}_Y$ are derived from $\mathbf{K}_X$ and $\mathbf{K}_Y$, respectively by setting their diagonal entries to zero: $\tilde{\mathbf{K}}_{X,ij} = (1-\delta_{ij})\mathbf{K}_{X,ij}$, where $\delta_{ij}$ is the Kronecker delta  (1 if $i=j$, and 0 otherwise). This estimator requires $n \ge 4$ for its standard definition.

%% file: files/method.tex
\section{\method: Hallucination Detection via Decoupled Representations}
\label{sec:irid}

Building upon the theoretical foundations of HSIC, as introduced in \cref{sec:foundations_hsic}, we propose \method, a single-pass approach for detecting hallucinations in LMs by investigating the internal model representations for input and output. \method~is based on the premise that hallucinations manifest as a quantifiable statistical decoupling between the LM's internal representations of the input sequence and the generated output.
We hypothesize that:
\begin{itemize}
    \item[\textbf{(i)}] \textbf{Sound Generation:} When an LM generates content that is grounded in the input context or in its parametric knowledge, its internal representation of the output, $H_{out}^{(\ell)}$, remains strongly statistically dependent on the internal representation of the input, $H_{in}^{(\ell)}$, at a given layer $\ell$. This strong coupling should result in a substantially positive HSIC value.
    
    \item[\textbf{(ii)}] \textbf{Hallucinated Generation:} Conversely, when an LM hallucinates, the generative process decouples the output representation $H_{out}^{(\ell)}$ from the input representation $H_{in}^{(\ell)}$. As the generated content diverges, these representations become increasingly independent. In such cases, their joint distribution approaches the product of their marginals, leading to an HSIC value close to zero, i.e., $\operatorname{HSIC}(H_{in}^{(\ell)}, H_{out}^{(\ell)}) \approx 0$.
\end{itemize}
\method~leverages this hypothesized drop in HSIC to detect hallucinations. A sufficiently low empirical HSIC score between selected input and output representations suggests a potential hallucination.

\subsection{Problem Formulation}
\label{subsec:irid_problem_formulation}
As defined previously in \cref{sec:background}, $\mathbf{S}_{in} = \{s_1, s_2, \ldots, s_I\}$ is an input token sequence of length $I$, and $\mathbf{S}_{out} = \{t_1, t_2, \ldots, t_O\}$ is the corresponding LM-generated output sequence of length $O$. For a chosen layer $\ell$, the matrix of hidden states for the input sequence is $\mathbf{H}_{in}^{(\ell)} \in \mathbb{R}^{I \times d}$, where $\mathbf{h}_{i,in}^{(\ell)} \in \mathbb{R}^d$ is the hidden state for input token $s_i$, and $d$ is the hidden state dimensionality. Similarly, for the output sequence, $\mathbf{H}_{out}^{(\ell)} \in \mathbb{R}^{O \times d}$, with rows $\mathbf{h}_{j,out}^{(\ell)} \in \mathbb{R}^d$.

To apply HSIC, we define random variables $X$ and $Y$ based on these representations:
\begin{definition}
\label{def:irid_random_variables}
\textit{Let $\mathcal{T}_{in}$ be the set of all possible token types that can be selected from an input sequence, and $\mathcal{T}_{out}$ be the set of all possible token types from an output sequence. We define two random variables, $X$ and $Y$, as :
\begin{itemize}
    \item $X: \mathcal{T}_{in} \to \mathbb{R}^d$, where for any selected input token type $\sigma_{in} \in \mathcal{T}_{in}$ (which, in a specific instance, corresponds to a token $s_i \in \mathbf{S}_{in}$), $X(\sigma_{in}) = \mathbf{h}_{i,in}^{(\ell)}$. The codomain of $X$ is $\mathcal{X} = \mathbb{R}^d$.
    \item $Y: \mathcal{T}_{out} \to \mathbb{R}^d$, where for any selected output token type $\sigma_{out} \in \mathcal{T}_{out}$ (which, in a specific instance, corresponds to a token $t_j \in \mathbf{S}_{out}$), $Y(\sigma_{out}) = \mathbf{h}_{j,out}^{(\ell)}$. The codomain of $Y$ is $\mathcal{Y} = \mathbb{R}^d$.
\end{itemize}}
\textit{For a single input-output pair $(\mathbf{S}_{in}, \mathbf{S}_{out})$, we independently select $n_{\text{eff}}$ tokens from $\mathbf{S}_{in}$ and $\mathbf{S}_{out}$, respectively, to obtain samples $\{x_1, \dots, x_{n_{\text{eff}}}\}$ for $X$ and $\{y_1, \dots, y_{n_{\text{eff}}}\}$ for $Y$. These form the two sets of observations for calculating empirical HSIC.}
\end{definition}

\subsection{\method~Score Computation}
\label{subsec:irid_algorithm_score}
The \method~framework for detecting hallucination in a single generation pass is summarized in Algorithm~\ref{alg:hide}. A key component of the proposed method is the computation of our HSIC-based score.

The standard unbiased HSIC estimator, $\widehat{\operatorname{HSIC}}_u$ (c.f. Equation~\ref{eq:unbiased_HSIC_foundation}), provides desirable statistical properties but is formally defined for $n \ge 4$ samples. In practical scenarios, particularly when processing short input or output sequences, the effective number of selected tokens, $n_{\text{eff}}$, may fall below this threshold. To ensure the applicability of our framework across all sequence lengths and maintain numerical stability, we employ a pragmatically adapted HSIC-like score, denoted by $\widehat{\operatorname{HSIC}}_{\text{\method}}$. This adaptation modifies the denominators in the standard unbiased HSIC estimator (Equation~\ref{eq:unbiased_HSIC_foundation}) by replacing terms of the form $(n - c)$ with $n$, and substituting the overall scaling factor $1/(n(n - 3))$ with $1/n^2$.

\begin{definition}
\label{def:hsic_irid_modified}
\textit{For an effective sample size $n \ge 1$, the $\widehat{\operatorname{HSIC}}_{\text{\method}}$ score is computed as:}
\begin{multline}
\widehat{\operatorname{HSIC}}_{\text{\method}}(X,Y) = \frac{1}{n^2} \bigg[ 
\text{Tr}(\tilde{\mathbf{K}}_X\tilde{\mathbf{K}}_Y) +
\frac{(\mathbf{1}^T \tilde{\mathbf{K}}_X \mathbf{1}) (\mathbf{1}^T \tilde{\mathbf{K}}_Y \mathbf{1})}{n^2} - \frac{2}{n} \mathbf{1}^T \tilde{\mathbf{K}}_X\tilde{\mathbf{K}}_Y \mathbf{1} \bigg]
\label{eq:hide-score}
\end{multline}
\textit{where $\tilde{\mathbf{K}}_X, \tilde{\mathbf{K}}_Y$ are the $n \times n$ Gram matrices with zeroed diagonals, and $\mathbf{1}$ is an $n \times 1$ vector of ones.}
\end{definition}
While this adaptation introduces bias relative to the standard $\widehat{\operatorname{HSIC}}_u$, it yields a score that is well-defined and numerically stable for any $n \ge 1$. This practical modification allows for consistent application of our method across all encountered data instances. We posit that $\widehat{\operatorname{HSIC}}_{\text{\method}}$ still serves as a valuable and usable heuristic for quantifying the relative strength of dependence, demonstrating its empirical effectiveness in detecting hallucinations in ~\cref{sec:results}.

\begin{algorithm}[ht!]
\caption{\textbf{\method}: Hallucination Detection via Decoupled Representations}
\label{alg:hide}
\begin{algorithmic}[1]
\Statex \textbf{Input:} Input sequence $\mathbf{S}_{in}$, Language model $M$, Layer index $\ell$, Token budget $n_{\text{eff}}$, RBF kernel bandwidth $\gamma$, Threshold $\tau$
\Statex \textbf{Output:} Decision: Hallucination or Non-hallucination

\Procedure{DetectHallucination}{$\mathbf{S}_{in}, M, \ell, n_{\text{eff}}, \gamma, \tau$}
    \State Initialize storage:
        \Statex \hspace{1em} $\mathbf{H}_{\mathrm{in}}^{(\ell)} \leftarrow [\,]$, \quad $\mathbf{H}_{\mathrm{out}}^{(\ell)} \leftarrow [\,]$
    \State Feed input $\mathbf{S}_{in}$ and auto-regressively generate output $\mathbf{S}_{out}$:
    \For{each token $t$ generated by $M$ during inference}
        \State Compute hidden states at layer $\ell$
        \If{$t \in \mathbf{S}_{in}$}
            \State Append its hidden representation to $\mathbf{H}_{\mathrm{in}}^{(\ell)}$
        \Else
            \State Append to $\mathbf{H}_{\mathrm{out}}^{(\ell)}$
        \EndIf
    \EndFor
    \State Now $\mathbf{H}_{\mathrm{in}}^{(\ell)}, \mathbf{H}_{\mathrm{out}}^{(\ell)}$ hold all relevant states

    \State $n_{\text{eff}} \gets \min(n_{\text{eff}}, |\mathbf{S}_{in}|, |\mathbf{S}_{out}|)$
    \If{$n_{\text{eff}} < 1$}
        \State \Return Undetermined
    \EndIf

    \State Identify top-$n_{\text{eff}}$ key-tokens:
        \Statex \hspace{1em} $\mathcal{K}_{\mathrm{in}} \leftarrow \text{KeyBERT}(\mathbf{S}_{\mathrm{in}}, n_{\text{eff}})$
        \Statex \hspace{1em} $\mathcal{K}_{\mathrm{out}} \leftarrow \text{KeyBERT}(\mathbf{S}_{\mathrm{out}}, n_{\text{eff}})$

    \State Collect hidden vectors for selected tokens:
        \Statex \hspace{1em} $\mathbf{X} \leftarrow \{\mathbf{h}_{i,\mathrm{in}}^{(\ell)} \mid \text{token}_i \in \mathcal{K}_{\mathrm{in}}\}$
        \Statex \hspace{1em} $\mathbf{Y} \leftarrow \{\mathbf{h}_{j,\mathrm{out}}^{(\ell)} \mid \text{token}_j \in \mathcal{K}_{\mathrm{out}}\}$

    \State Build RBF kernel $k_{\mathrm{rbf}}$ with bandwidth $\gamma$
    \State Compute HIDE score:
        \Statex \hspace{1em} $\text{score} \gets \widehat{\mathrm{HSIC}}_{\mathrm{HIDE}}(\mathbf{X}, \mathbf{Y}; k_{\mathrm{rbf}})$

    \If{$\text{score} < \tau$}
        \State \Return Hallucination
    \Else
        \State \Return Non-hallucination
    \EndIf
\EndProcedure
\end{algorithmic}
\end{algorithm}

\subsection{Properties of the \method~Score}
\label{sec:properties-hide}
We now analyze the properties of our proposed \method~score (Equation ~\ref{eq:hide-score}), denoted as \(\widehat{\operatorname{HSIC}}_{\method}\), which adapts the standard unbiased V-statistic based HSIC estimator to be computable for any number of effective tokens \(n \equiv n_{\text{eff}} \ge 1\). 

Let \(\{(x_i,y_i)\}_{i=1}^{n}\) be the set of hidden state representations for the selected input and output tokens. We use the RBF kernel, which is bounded such that \(0 < k(u,v) \le 1\) for any vectors \(u,v\). The score is computed using Gram matrices with zeroed diagonals, \(\tilde{\mathbf{K}}_X\) and \(\tilde{\mathbf{K}}_Y\), as defined in Equation~\ref{eq:unbiased_HSIC_foundation}.

\begin{lemma}[\textbf{Behaviour for Small Sample Sizes}]
\label{lem:small_n_properties}
\textit{The \(\widehat{\operatorname{HSIC}}_{\method}\) estimator is computable for small sample sizes where \(\widehat{\operatorname{HSIC}}_u\) is undefined. Specifically:}
\begin{enumerate}
    \item[\textit{(i)}] \textit{If \(n=1\), then \(\widehat{\operatorname{HSIC}}_{\method}(X,Y) = 0\).}
    \item[\textit{(ii)}] \textit{If \(n=2\), then \(\widehat{\operatorname{HSIC}}_{\method}(X,Y) = \frac{1}{4} k_X(x_1,x_2)k_Y(y_1,y_2)\), which implies \(0 < \widehat{\operatorname{HSIC}}_{\method}(X,Y) \le 1/4\).}
\end{enumerate}
\end{lemma}

\begin{lemma}[\textbf{Asymptotic Properties}]
\label{lem:asymptotic_properties}
\textit{The \(\widehat{\operatorname{HSIC}}_{\method}\) estimator is asymptotically unbiased and strongly consistent.}
\begin{enumerate}
    \item[\textit{(i)}] \textit{(Asymptotic Bias) The bias of the estimator vanishes as \(n \to \infty\):
    \[ \bigl|\mathbb{E}[\widehat{\operatorname{HSIC}}_{\method}] - \operatorname{HSIC}(X,Y)\bigr| = O(1/n). \]}
    \item[\textit{(ii)}] \textit{(Strong Consistency) As \(n \to \infty\), \(\widehat{\operatorname{HSIC}}_{\method}(X,Y)\) converges almost surely to the true population HSIC:
    \[ \widehat{\operatorname{HSIC}}_{\method}(X,Y) \xrightarrow{\text{a.s.}} \operatorname{HSIC}(X,Y). \]}
\end{enumerate}
\end{lemma}

The detailed proofs of the above lemmas are discussed in Appendix A.

\subsection{Methodological Choices}
\label{subsec:irid_choices}

Our design involves two key choices that must be fixed before the
\method~score can be evaluated:  
(\textit{i}) how to align the \emph{token sets} drawn from the input-output
sequences so that the \emph{same} number of samples are drawn for both random variables $X$ and $Y$, and
(\textit{ii}) from which  \emph{decoder layer}, $\ell\!\in\!\{1,\dots,L\}$ the
hidden-state matrices $\mathbf H^{(\ell)}_{\text{in}}$ and
$\mathbf H^{(\ell)}_{\text{out}}$ are taken.  We formalise the two
choices below and motivate the instantiations that are used throughout the
paper.

\subsubsection{Strategy for Aligning the Sets of Input and Output Samples}
\label{sec:token_alignment}
Let $\mathbf H^{(\ell)}_{\text{in}}\!\in\!\mathbb R^{I\times d}$ and
$\mathbf H^{(\ell)}_{\text{out}}\!\in\!\mathbb R^{O\times d}$ be the matrices
of hidden states for the input and output sequences, respectively, at layer $\ell$, where $I$ and $O$ may differ by in magnitude.  
To guarantee that $X$ and $Y$ are constructed from the \emph{same} effective
sample size, $n_{\text{eff}}$, we consider two principled strategies:

\begin{enumerate}[leftmargin=1.2em,label=(\roman*)]
\item\textbf{Keyword-based Subsampling.}\;
  Let $n_{\text{eff}}$ be our \emph{token budget}, i.e., the number of representative tokens we want to select from both the input and output. Since, in some cases, the input or the output can be smaller than our token budget, we set the value of $n_{\text{eff}}$ to be the minimum of our chosen token budget, input length and output length. We then extract a set of $n_{\text{eff}}$ salient tokens
  $\mathcal K_{\text{in}}\subseteq \mathbf{S}_{\text{in}}$ and
  $\mathcal K_{\text{out}}\subseteq \mathbf{S}_{\text{out}}$ using KeyBERT, which acts as a maximal-marginal-relevance ranker. 
  The aligned sample sets are then
  $\mathcal X=\{\,\mathbf h^{(\ell)}_{i,\text{in}}\mid s_i\!\in\!\mathcal
  K_{\text{in}}^{(n_{\text{eff}})}\}$ and
  $\mathcal Y=\{\,\mathbf h^{(\ell)}_{j,\text{out}}\mid
  t_j\!\in\!\mathcal K_{\text{out}}^{(n_{\text{eff}})}\}$, where the superscript
  $n_{\text{eff}}$ denotes truncation to the top-$n_{\text{eff}}$ representative tokens.  In practice we
  use $n_{\text{eff}}=20$ for our experiments, unless explicitly specified otherwise. We discuss more on the effect of varying our token budget in~\cref{sec:token_budget}. 

\item\textbf{Rank-truncated SVD Alignment.}\;
  Alternatively, we avoid lexical heuristics altogether and enforce equal
  sample size by projecting \emph{both} matrices $\mathbf H^{(\ell)}_{\text{in}}$ and $\mathbf H^{(\ell)}_{\text{out}}$ to subspaces with the same rank (i.e., $n_{\text{eff}})$.
  Concretely, for $\mathbf H^{(\ell)}_{\text{in}}$ we compute a thin SVD
  $\mathbf H^{(\ell)}_{\text{in}}
     =\mathbf U_{X}\,\boldsymbol\Sigma_{X}\mathbf V_{X}^{\top}$ and retain the
  $n_{\text{eff}}$ largest singular directions,
  \begin{equation}
     \mathbf X_{r}
        =\boldsymbol\Sigma_{X}^{(n_{\text{eff}})}\,
         \bigl(\mathbf V_{X}^{(n_{\text{eff}})}\bigr)^{\!\top}\in\mathbb R^{n_{\text{eff}}\times d}.
  \end{equation}
  An analogous projection $\mathbf Y_{r}$ is obtained for
  $\mathbf H^{(\ell)}_{\text{out}}$, yielding aligned sample sets
  $\mathcal X=\{\mathbf x_{r}^{(1)},\dots,\mathbf x_{r}^{(n_{\text{eff}})}\}$ and
  $\mathcal Y=\{\mathbf y_{r}^{(1)},\dots,\mathbf y_{r}^{(n_{\text{eff}})}\}$.
  As both $\mathbf X_{r}$ and $\mathbf Y_{r}$ now contain exactly $n_{\text{eff}}$
  \emph{orthogonal} rows, the empirical Gram matrices in Equation~\eqref{eq:hide-score} have identical dimensions, eliminating the need for any further padding or truncation. We discuss the performances of \method~with both these alignment strategies in~\cref{sec:token_selection}. Unless stated otherwise, we adopt the keyword-based subsampling strategy for our experiments.
\end{enumerate}

\subsubsection{Layer Selection}
Each decoder layer captures different levels of abstraction; therefore, the choice of $\ell$ can, in principle, influence the measured dependence. Based on prior work suggesting that representations corresponding to the middle layers are often richer \citep{skean2025layerlayeruncoveringhidden} and provide stronger sentence-level semantic information suitable for detecting inconsistencies \citep{azaria2023internalstatellmknows}, we fix $\ell=\ell_{\text{mid}}$ for all our experiments to avoid an additional hyperparameter sweep. However, our framework is flexible to any layer choice and we observe the performance of \method~to be almost agnostic to the choice of $\ell$ (\cref{sec:layer_choice}).

%% file: files/results_table.tex
\afterpage{%
  \clearpage
\begin{table*}[p]
\centering
\caption{AUC-ROC and PCC values using sentence similarity and ROUGE-L as correctness measures for faithfulness and factuality hallucination detection across four datasets. \method~is compared with single-pass and multi-pass baselines (rows corresponding to multi-pass are highlighted in gray; five outputs are generated for each multi-pass method). The best result for each metric is shown in \textbf{bold}; the best among single-pass methods is \underline{underlined}.}

\rotatebox{90}{
\adjustbox{max width=0.9\textheight, max height=\linewidth, keepaspectratio}{
\begin{tabular}{lcccc|cccc|cccc|cccc|cccc}
\toprule
\multirow{3}{*}{Method} 
& \multicolumn{8}{c}{\textbf{Faithfulness}} 
& \multicolumn{8}{c}{\textbf{Factuality}} \\
\cmidrule(lr){2-9} \cmidrule(lr){10-17}
& \multicolumn{4}{c}{RACE} & \multicolumn{4}{c}{SQuAD} 
& \multicolumn{4}{c}{NQ} & \multicolumn{4}{c}{TriviaQA} & \multicolumn{4}{c}{\textbf{Average}}\\
\cmidrule(lr){2-5} \cmidrule(lr){6-9} \cmidrule(lr){10-13} \cmidrule(lr){14-17} \cmidrule(lr){18-21}
& $\text{AUC}_\text{s}$ & $\text{AUC}_\text{r}$ & $\text{PCC}_\text{s}$ & $\text{PCC}_\text{r}$ 
& $\text{AUC}_\text{s}$ & $\text{AUC}_\text{r}$ & $\text{PCC}_\text{s}$ & $\text{PCC}_\text{r}$ 
& $\text{AUC}_\text{s}$ & $\text{AUC}_\text{r}$ & $\text{PCC}_\text{s}$ & $\text{PCC}_\text{r}$ 
& $\text{AUC}_\text{s}$ & $\text{AUC}_\text{r}$ & $\text{PCC}_\text{s}$ & $\text{PCC}_\text{r}$
& $\text{AUC}_\text{s}$ & $\text{AUC}_\text{r}$ & $\text{PCC}_\text{s}$ & $\text{PCC}_\text{r}$ \\
\hline
\multicolumn{21}{c}{\cellcolor[HTML]{C0C0C0}Llama-3.2-3B} \\\hline
Perplexity         & 50.27 & 49.94 & \underline{18.6} & 15 & 46.53 & 47.3 & -7.81 & -1.56 & 53.03 & 54.25 & -8.16 & 9.2 &57.53 & 57.01 & 6.14 & \underline{20.41} & 51.84 & 52.13 & 2.19 & 10.76 \\

Energy             & 49.07 & 49.93 & 13.10 & 9.14 & 36.9 & 37.22 & -29.49 & -26.22 & 36.83 & 35.2 & -25.81 & -10.75 &43.32 & 41.84 & -34.53 & -22.20 & 41.53 & 41.05 & -19.18 & -12.51 \\

\rowcolor{gray!20} LN-Entropy & 50.90 & 50.26 & 6.01 & 6.96 & 60.85 & 60.89 & 14.24 & 19.74 & 72.39 & 71.31 & 20.58 & 22.53 & 67.31 & 67.24 & 22.52 & 28.82 & 62.86 & 62.42 & 15.84 & 19.52 \\
\rowcolor{gray!20} Lexical Similarity & 64.11 & 65.04 & \textbf{22.04} & \textbf{27.32} & 64.61 & 66.71 & 28.81 & 34.1 & 65.98 & 69.51 & 20.32 & 27.8 &  \textbf{79.74} & \textbf{81.34} & \textbf{52.99}& \textbf{56.59} & 68.61 & \textbf{70.65} & 31.04 & \textbf{36.45}\\
\rowcolor{gray!20} Eigenscore         & 58.17 & 57.18 & 9.56 & 12.37 & 65.31 & 65.49 & 21.29 & 25.46 & 68.34 & 67.34 & 20.27 & 17.39 & 76.13 & 76.57 & 50.45 & 48.15 & 66.98 & 66.65 & 25.39 & 25.85 \\
\textbf{\method} & \textbf{\underline{66.43}} & \textbf{\underline{65.48}} & {18.34} & \underline{20.72} & \textbf{\underline{77.55}} & \textbf{\underline{77.28}} & \textbf{\underline{49.86}} & \textbf{\underline{50.3}} & \textbf{\underline{75.16}} & \textbf{\underline{74.63}} & \textbf{\underline{47.77}}& \textbf{\underline{25.77}} & \underline{58.36} & \underline{59.14} & \underline{27.46} & 17.88 & \textbf{\underline{69.37}} & \underline{69.13} & \textbf{\underline{35.86}} & \underline{28.67} \\
\hline
\multicolumn{21}{c}{\cellcolor[HTML]{C0C0C0}Llama-3.2-3B-Instruct} \\\hline
Perplexity         & 59.52 & 61.28 & 34.59& 28.44 & 52.50 & 55.22 & 9.73 & 15.36 & 54.90 & 54.16 & -7.26 & 8.80& \underline{60.08} & \underline{60.14} & 6.42 & \underline{22.49} & 56.75 & 57.70 & 10.87 & 18.77\\
Energy             & 54.19 & 57.03 & 21.48 & 17.04 & 37.36 & 38.57 & -18.23 & -16.81 & 25.89 & 24.72 & -48.16 & -25.99& 42.51 & 41.59 & -31.54 & -16.17 & 39.99 & 40.48 & -19.11 & -10.48\\
\rowcolor{gray!20} \rowcolor{gray!20} LN-Entropy         & 54.46 & 55.65 & 14.96 & 12.96 & 62.98 & 64.74 & 23.51 & 29.69 & 71.88 & 70.42 & 25.68& 24.93& 71.83 & 72.01 & 22.95 & 29.18 & 65.29 & 65.70 & 21.77 & 24.19\\
\rowcolor{gray!20} Lexical Similarity & 66.75 & 68.20 & 31.91 & 33.83 & 69.05 & 71.69 & 35.08 & 41.91 & 70.49 & 70.53& 34.54 & 32.62& \textbf{79.25} & \textbf{80.59} & 51.98 & \textbf{54.42} & 71.38 & \textbf{72.75} & 38.38 & \textbf{40.69} \\
\rowcolor{gray!20} Eigenscore         & 62.82 & 62.09 & 19.09 & 20.92 & 73.14 & 73.57 & 39.58 & \textbf{45.56} & 76.42 & 75.56& 41.64 & 32.71& 78.35 & 79.08 & \textbf{53.46} & 49.55 & \textbf{72.68} & 72.57 & 38.44 & 37.19 \\
\textbf{\method} & \textbf{\underline{74.88}} & \textbf{\underline{73.02}} & \textbf{\underline{27.36}} & \textbf{\underline{33.95}} & \textbf{\underline{75.18}} & \textbf{\underline{75.48}} & \textbf{\underline{42.80}} & \underline{44.98} & \textbf{\underline{76.56}} & \textbf{\underline{77.27}} & \textbf{\underline{53.91}} & \textbf{\underline{35.40}} &58.43 & 59.03 & \underline{30.64} & 19.54 & \underline{71.26} & \underline{71.20} & \textbf{\underline{38.68}} & \underline{33.47}\\
\hline
\multicolumn{21}{c}{\cellcolor[HTML]{C0C0C0}Llama-3-8B} \\\hline
Perplexity         & 50.78 & 51.93 & \underline{23.01} & 18.07 & 45.08 & 46.37 & -8.01 & -2.39 & 48.18 & 47.51 & -11.22 & 2.16& 38.21 & 36.90 & -19.49 & -8.54 & 45.56 & 45.68 & -3.92 & 2.33 \\
Energy             & 50.10 & 51.35 & 17.68& 12.31 & 35.27 & 35.48 & -29.16 & -26.95 & 33.70 & 32.13 & -23.86 & -15.02&27.43 & 26.48& -50.84 & -42.86 & 36.63 & 36.36 & -21.55 & -18.13 \\
\rowcolor{gray!20} LN-Entropy         & 45.49 & 45.85 & -0.10 & -0.54 & 60.50 & 60.57 & 15.09& 19.82 & 65.81 & 65.00 & 13.80 & 16.17  & 72.15& 72.08 & 33.54 & 37.79 & 60.99 & 60.87 & 15.58 & 18.31  \\
\rowcolor{gray!20} Lexical Similarity & 66.10 & 66.92 & \textbf{29.15} & \textbf{33.80} & 64.01 & 66.57 & 29.36 & 33.69 & 65.22 & 65.84 & 20.10 & 25.87& 77.13 & 78.65 & 47.18 & 50.24 & 68.12 & 69.50 & 31.45 & \textbf{35.90} \\
\rowcolor{gray!20} Eigenscore         & 54.16 & 53.98 & 6.73& 10.88 & 66.57 & 66.34 & 25.66 & 29.93 & 64.36 & 64.76 & 17.07 & 16.90& \textbf{81.06}& \textbf{82.14} & \textbf{57.92} & \textbf{57.46} & 66.54 & 66.81 & 26.84 & 28.79 \\
\textbf{\method} & \textbf{\underline{69.81}} & \textbf{\underline{66.95}} & 16.76 & \underline{22.87} & \textbf{\underline{76.87}} & \textbf{\underline{77.45}} & \textbf{\underline{49.09}} & \textbf{\underline{50.43}} & \textbf{\underline{79.02}} & \textbf{\underline{79.17}} & \textbf{\underline{49.61}} & \textbf{\underline{33.91}} & \underline{65.65} & \underline{66.82} & \underline{41.82} & \underline{35.19} & \textbf{\underline{72.84}} & \textbf{\underline{72.60}} & \textbf{\underline{39.32}} & \underline{35.60}\\
\hline
\multicolumn{21}{c}{\cellcolor[HTML]{C0C0C0}Llama-3-8B-Instruct} \\\hline
Perplexity         & 60.20 & 61.91 & \underline{22.52} & 22.42 & 47.46 & 50.73 & 2.95 & 6.94 & 57.47 & 56.11 & -7.64 & 9.72& \underline{61.70} & 61.14 & 7.68 & 21.43 & 56.71 & 57.47 & 6.38& 15.13\\
Energy             & 55.94 & 57.16 & 18.26 & 15.97 & 43.05 & 44.60 & -8.96 & -8.31 & 30.63 & 29.56 & -36.02 & -18.50 &39.40 & 38.80 & -29.22 & -16.69 & 42.25 & 42.53 & -13.99 & -6.88 \\
\rowcolor{gray!20} LN-Entropy         & 57.13 & 57.83 & 12.59 & 14.26 & 62.73 & 63.44 & 23.11 & 26.94 & 67.29 & 65.43 & 18.12 & 20.80 & 79.61 & 79.72 & 42.76 & 48.41 & 66.69 & 66.60 & 24.15 & 27.60 \\
\rowcolor{gray!20} Lexical Similarity & 69.28 & 71.22 & \textbf{36.64} & \textbf{40.07} & 62.46 & 65.14 & 25.76 & 29.69 & 69.05 & 68.88 & 26.24 & 28.84&81.12 & 82.33 & 58.68& 59.19 & 70.48 & 71.89 & 36.83 & 39.45\\
\rowcolor{gray!20} Eigenscore         & 63.65 & 62.99 & 18.78& 23.01 & 72.11 & 71.49 & 36.95 & 39.85 & 77.89 & 77.53 & 41.23 & 32.73& \textbf{82.83} & \textbf{83.94} & \textbf{69.32} & \textbf{63.72} & \textbf{74.12} & \textbf{73.99} & 41.57 & \textbf{39.83} \\
\textbf{\method} & \textbf{\underline{73.20}} & \textbf{\underline{72.73}} & 22.12 & \underline{30.11} & \textbf{\underline{80.15}} & \textbf{\underline{79.14}} & \textbf{\underline{52.27}} & \textbf{\underline{55.98}} & \textbf{\underline{79.29}} & \textbf{\underline{78.91}} & \textbf{\underline{56.33}} & \textbf{\underline{37.98}} & \underline{61.70} & \underline{62.94} & \underline{36.73} & \underline{27.27} & \underline{73.58} & \underline{73.43} & \textbf{\underline{41.86}} & \underline{37.84}\\
\hline
\multicolumn{21}{c}{\cellcolor[HTML]{C0C0C0}Gemma-2-9B} \\\hline
Perplexity         & \underline{59.88} & \underline{60.65} & \underline{29.89} & \underline{27.98} & 47.42 & 48.29 & -8.61 & -1.85 & 29.92 & 35.59 & 9.29 & 12.95 & 43.08 & 46.18 & -6.69 & 6.65 & 45.07 & 47.68 & 5.97 & 11.43 \\
Energy             & 44.05 & 43.20 & -13.38 & -12.32 & 52.12 & 52.50 & 6.81 & 2.25 & 64.89 & 60.48 & 0.50 & -3.93& 68.90 & 67.80 & 14.82 & 7.61 & 57.49 & 56.00 & 2.19 & -1.60\\
\rowcolor{gray!20} LN-Entropy         & 63.27 & 64.11 & 25.93 & 26.76& 60.05 & 61.26 & 15.72 & 22.29 & 44.14 & 53.01 & 8.19 & 20.20 & 41.53 & 44.85 & 0.95 & 13.07 & 52.25 & 55.81 & 12.70 & 20.58 \\
\rowcolor{gray!20} Lexical Similarity & \textbf{69.02} & \textbf{69.62} & \textbf{34.18} & \textbf{35.48} & 67.10 & 70.32 & 35.25 & 40.56 & 56.80 & 59.43 & 13.50 & \textbf{23.50} &62.98 & 65.46 & 23.79 & 32.99 & 63.97 & 66.21 & 26.68 & \textbf{33.13}\\
\rowcolor{gray!20} Eigenscore         & 64.58 & 64.35 & 24.99 & 25.20 & 73.40 & 75.14 & 41.66 & 47.29 & 73.31 & 74.31 & 7.70 & 16.79 & 71.40 & 71.70 & 23.15 & 27.42 & 70.67 & 71.37 & 24.37 & 29.18\\
\textbf{\method} & 58.06 & 57.99 & 20.60 & 16.50 & \textbf{\underline{75.17}} & \textbf{\underline{76.66}} & \textbf{\underline{47.63}} & \textbf{\underline{48.43}} & \textbf{\underline{79.08}} & \textbf{\underline{75.96}} & \textbf{\underline{24.08}} & \underline{15.41} & \textbf{\underline{80.84}} & \textbf{\underline{81.02}} & \textbf{\underline{50.51}} & \textbf{\underline{43.88}} & \textbf{\underline{73.29}} & \textbf{\underline{72.91}} & \textbf{\underline{35.70}} & \underline{31.05} \\ 
\hline
\multicolumn{21}{c}{\cellcolor[HTML]{C0C0C0}Gemma-2-9B-Instruct} \\\hline
Perplexity         & 60.22& 61.27 & \underline{14.11} & \underline{17.12} & 53.15 & 56.90 & 5.73 & 12.49 & 31.52 & 34.14 & 29.54 & 0.19 & 32.09 & 36.85 & 52.60 & 11.92 & 44.25 & 47.29 & 25.49 & 10.43\\
Energy             & 57.29 & 55.87 & 7.28 & 5.99 & \textbf{\underline{80.08}} & \textbf{\underline{77.41}} & \textbf{\underline{49.54}} & \textbf{\underline{50.08}} & 77.19 & 76.56 & -9.51 & 11.55 & \textbf{\underline{92.34}} & \textbf{\underline{90.96}} & -12.88 & 28.05 & \textbf{\underline{76.73}} & \textbf{\underline{75.20}} & 8.61 & 23.92\\
\rowcolor{gray!20} LN-Entropy         & 62.82 & 64.58 & 17.67 & 20.84 & 67.09 & 69.50 & 26.65 & 33.47 & 44.68 & 46.01 & 13.79 & 4.64 & 39.70 & 44.13 & 28.49 & 14.62 & 53.57 & 56.06 & 21.65 & 18.39 \\
\rowcolor{gray!20} Lexical Similarity & \textbf{72.55} & \textbf{72.98} & \textbf{34.16} & \textbf{38.63} & 72.33 & 74.97 & 39.77 & 46.57 & 53.27 & 56.91 & 19.80 & 11.13 & 51.56 & 56.22 & 31.36 & 21.47 & 62.43 & 65.27 & 31.27 & 29.45  \\
\rowcolor{gray!20} Eigenscore         & 70.22 & 70.59 & 24.78 & 29.44 & 71.70 & 73.99 & 44.71 & 49.87 & 65.97 & 67.11 & 8.40 & 14.46 & 61.17 & 64.63 & 15.05 & 21.13 & 67.26 & 69.08 & 23.24 & 28.73 \\
\textbf{\method} & \underline{64.18} & \underline{64.03} & 11.92 & 16.26 & 72.69 & 73.32 & 39.73 & 39.70 & \textbf{\underline{85.19}} & \textbf{\underline{85.89}} & \textbf{\underline{51.20}} & \textbf{\underline{41.99}} & 70.90 & 70.72 & \textbf{\underline{52.64}} & \textbf{\underline{45.90}} & 73.24 & 73.49 & \textbf{\underline{38.87}} & \textbf{\underline{35.97}}\\
\midrule
\end{tabular}
}
}
\label{tab:results_exp}
\end{table*}
}

%% file: files/exp_setup.tex
\section{Experimental Setup}
\label{sec:experiments}

\paragraph{\textbf{Evaluation Datasets}}
Following the setup of \citet{hallucinations-leaderboard}, we evaluate \method~on four benchmark datasets, as discussed below, to assess its effectiveness in detecting both faithfulness and factuality hallucinations:
\begin{itemize}
    \item[\textbf{(i)}] \textbf{RACE (ReAding Comprehension dataset from Examinations)} \citep{lai-etal-2017-race} is a multiple-choice reading comprehension dataset from English exams for Chinese students, featuring 27,933 passages and 97,867 questions. We use the test split of the `high' subset containing 1,050 passages and 3,498 questions. RACE is challenging due to its focus on inference and reasoning. RACE is used to evaluate \textit{faithfulness hallucinations}, as answers must be supported by explicit information in the provided passage, making inference and grounding critical.
    
    \item[\textbf{(ii)}] \textbf{SQuAD (Stanford Question Answering Dataset)} \citep{rajpurkar2016squad100000questionsmachine} is a widely used benchmark for extractive question answering, containing crowd-sourced questions on Wikipedia articles. SQuAD-v1.1 contains only questions for which answers are present within the provided context, while SQuAD-v2.0 extends it to include both answerable and unanswerable questions. Following \citet{chen_inside:_2024}, we use only answerable questions from the development split of $\text{v}2.0$, retaining 5,928 context-question-answer pairs. SQuAD also targets \textit{faithfulness hallucinations}, as the correct answer must be a text span from the passage, requiring the model to remain faithful to the context.

    \item[\textbf{(iii)}] \textbf{NQ (Natural Questions)} \citep{kwiatkowski-etal-2019-natural} is a large-scale question answering dataset consisting of real, anonymized queries issued to the Google search engine. Following \citet{lin2024generatingconfidenceuncertaintyquantification}, we use the validation set consisting of 3610 pairs. NQ is notable for its use of naturally occurring user queries, making it a realistic and challenging benchmark for end-to-end QA. NQ is primarily used to assess \textit{factuality hallucinations}, as it requires models to provide correct answers to open-domain questions based on external world knowledge, testing factual correctness.

    \item[\textbf{(iv)}] \textbf{TriviaQA} \citep{joshi2017triviaqalargescaledistantly} is a large-scale reading comprehension dataset with over 95,000 crowd-authored question-answer pairs and associated evidence documents. The questions are naturally complex, often requiring cross-sentence reasoning. We use the \texttt{rc.nocontext} validation subset, which contains 9,960 deduplicated pairs. TriviaQA also focuses on \textit{factuality hallucinations}, as we use the subset which has no input context, hence the evaluation relies on accurate, fact-based answers.
\end{itemize}

\paragraph{\textbf{Models and Generation Configurations}} To ensure that our findings are broadly applicable and robust across diverse model scales and variants, we consider six open-source LMs, which include both \textit{base} and \textit{instruction-tuned} variants of Llama-3.2-3B, Llama-3-8B, and Gemma-2-9B. All model checkpoints are sourced directly from the HuggingFace model hub without any additional fine-tuning. For all experiments, we generate outputs in a \textit{zero-shot setting}, using greedy decoding for techniques requiring a single output, and set the values of temperature, top-$p$ and top-$k$ parameters to $0.5$, $0.99$ and $10$, respectively for stochastic sampling in case of methods requiring multiple outputs for the same input prompt. All our experiments were run on a single NVIDIA A100-SXM4-80GB GPU.

\paragraph{\textbf{Evaluation Metrics}}

To comprehensively assess the effectiveness of \method~and baseline methods in hallucination detection, we employ a variety of evaluation metrics commonly used in the literature. 
\begin{itemize}
    \item \textbf{Area Under the ROC Curve (AUC-ROC):} This threshold-independent metric quantifies how well a given method can distinguish between hallucinated and non-hallucinated outputs. Higher AUC-ROC values indicate stronger binary discrimination.
    \item \textbf{Pearson Correlation Coefficient (PCC):} PCC measures the linear correlation between the predicted scores of a method (e.g., \method~score) and the ground-truth correctness labels. This reflects the extent to which the method's confidence scores align with true instance-level correctness.
\end{itemize}
We report AUC-ROC and PCC values for both \emph{sentence embedding similarity} and \emph{ROUGE-L overlap} as the reference ground-truth measure, denoted in the results as $AUC_s$, $PCC_s$ (using sentence similarity) and $AUC_r$, $PCC_r$ (using ROUGE-L). The sentence similarity is calculated as the cosine similarity between the sentence embeddings of the generated output and the reference answer, using the \textit{nli-roberta-large} model~\citep{reimers-gurevych-2019-sentence} -- an output is considered correct if its similarity exceeds a fixed threshold set to $0.9$. When using the ROUGE-L measure~\citep{lin-2004-rouge}, outputs with an overlap score of more than $0.5$ with the reference are considered to be correct. We also report the AUC-ROC and PCC values for exact match with the ground-truth answer as the correctness measure.

%% file: files/results.tex
\section{Results and Discussions}
\label{sec:results}

We now present a thorough evaluation of \method~against both single-pass and multi-pass baseline methods, focusing on its ability to detect faithfulness and factuality hallucinations, as well as its computational efficiency. We further discuss notable trends and provide an assessment of areas where \method~is particularly effective, as well as scenarios where its advantages are less pronounced. The performance of \method~across four datasets and six models with $AUC_s$, $AUC_r$, $PCC_s$ and $PCC_r$ as evaluation metrics, compared against six baselines consisting of state-of-the-art single-pass and multi-pass detection techniques, are detailed in Table~\ref{tab:results_exp}. Further, the performances with AUC-ROC and PCC values calculated based on exact-match as the correctness measure are reported in Table~\ref{tab:exact_match} of Appendix B.

\subsection{Faithfulness Hallucination Detection}
\label{sec:faithful_results}

Faithfulness hallucinations, characterized by outputs unsupported by the provided context, pose a critical challenge for reliable QA systems. Our results on SQuAD and RACE, as summarized in Tables~\ref{tab:results_exp} and~\ref{tab:exact_match}, demonstrate that \method~consistently outperforms single-pass baselines, such as Perplexity and Energy-score, in $10$ out of $12$ $(model, dataset)$ combinations. Across all models on SQuAD and RACE, \method~achieves respective average relative improvements of $47.4\%$ and $20.1\%$ in $AUC_s$ scores over the best-performing single-pass baseline.

\method's advantage persists when compared to the state-of-the-art multi-pass methods. On SQuAD, for example, \method~achieves an $AUC_s$ score of $80.15\%$ with Llama-3-8B-Instruct, which is an absolute improvement of $8$ percentage points over Eigenscore, the state-of-the-art multi-pass baseline. The improvements are also reflected in PCC values, with \method~outperforming single-pass baselines and matching the best multi-pass alternatives in most settings. This trend is also mirrored on the RACE dataset, where \method~provides consistently robust performance across both base and instruction-tuned variants of all models. Overall, for faithfulness hallucination detection, \method~achieves an average relative gain of $33.7\%$ and $6.5\%$ in $AUC_s$ over the  single-pass and multi-pass baselines, respectively. These results collectively underscore the effectiveness of measuring internal representational dependence as a robust strategy for detecting faithfulness hallucinations across diverse language model architectures and question-answering scenarios.

\begin{figure}[t!]
    \centering
    \includegraphics[width=\columnwidth]{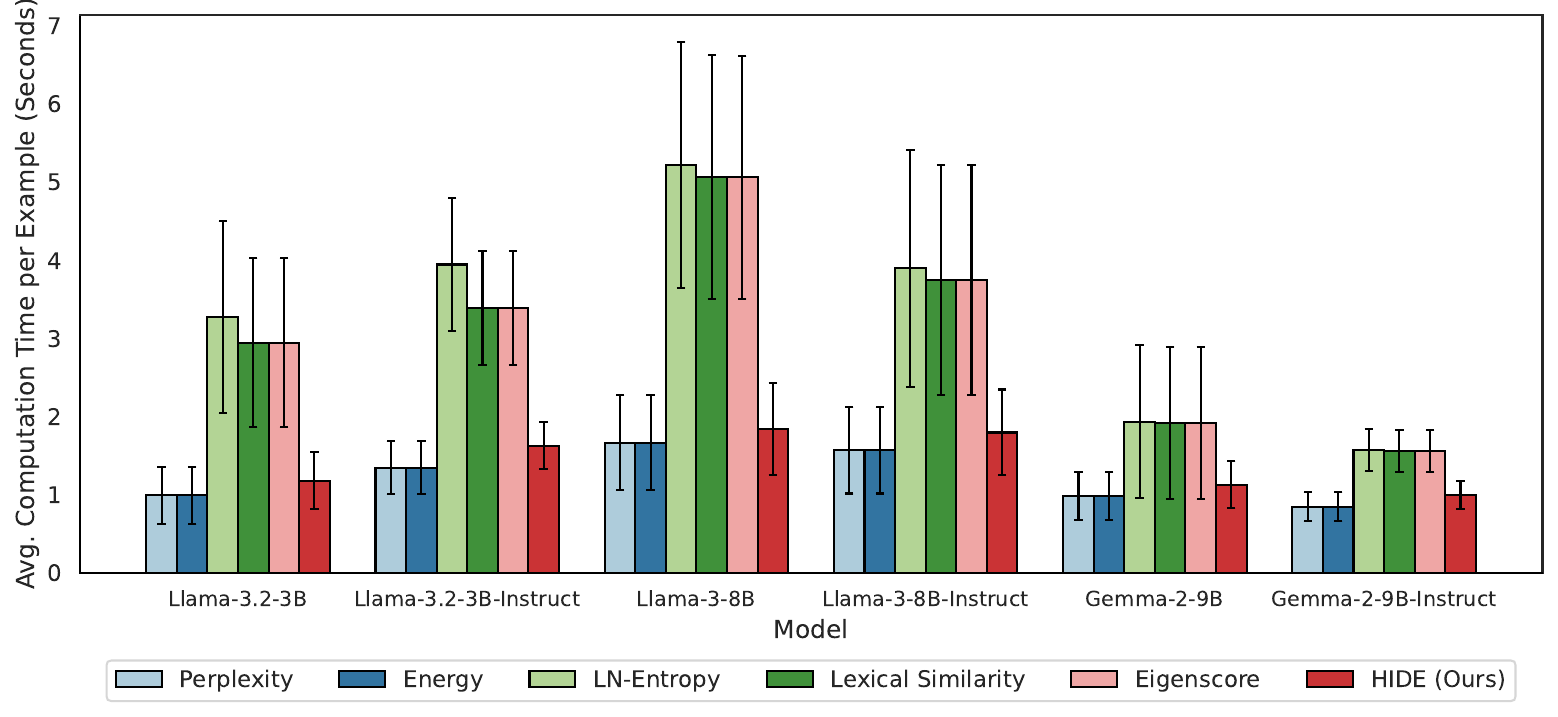}
    \caption{Comparison of the average computation time required by different hallucination detection methods to process each example, for six models. The averaging is done across all instances in the four datasets, i.e. RACE, SQuAD, NQ and TriviaQA. We observe that the average computation time required for \method~is similar to other single-pass methods and is $\sim 51\%$ less on average compared to the multi-pass methods generating $5$ outputs for each input.}
    \vspace{-4mm}
    \label{fig:time_analysis}
\end{figure}

\subsection{Factuality Hallucination Detection}
\label{sec:factual_results}

We further evaluate \method~on benchmarks designed to test factuality hallucinations, such as Natural Questions (NQ) and TriviaQA. Here, the outputs may be fluent yet factually unsupported. As shown in Tables~\ref{tab:results_exp} and~\ref{tab:exact_match}, even in the case of actuality hallucinations, \method~consistently provides improvements over single-pass baselines (in $10$ out of $12$ settings), with an average gain of $22.9\%$ in $AUC_s$ scores across models for NQ and TriviaQA. For instance, on NQ, \method~achieves $AUC_s$ score of $79.29\%$ with Llama-3-8B-Instruct, compared to $57.47\%$ for the best uncertainty-based single-pass competitor. The method also provides higher PCC values, indicating greater consistency with ground-truth correctness.

Compared to multi-pass methods, \method~is generally competitive and, in some settings, outperforms both Eigenscore and Lexical Similarity. For example, on the NQ dataset, \method~achieves higher AUC-ROC and PCC scores than all other methods for all six model variants. However, in certain configurations, such as TriviaQA with Llama-3-8B, the advantage of \method~is less pronounced, and Eigenscore achieves marginally higher AUC-ROC and PCC values. A closer examination of these cases suggests that datasets with extremely short or repetitive answers can reduce the discriminative power of representation-based dependence measures. We discuss such limitations in detail in \cref{sec:error_analysis}.

\subsection{Computational Efficiency}
\label{sec:compute_time}

An important goal of \method~is to balance detection accuracy with computational efficiency. As illustrated in Figure~\ref{fig:time_analysis}, the average time required for computation of the \method~score is similar to other single-pass methods. We observe that \method~reduces computational time by approximately $51\%$ on average compared to methods requiring multiple generations, i.e., LN-Entropy, Lexical Similarity, and Eigenscore (we generated $5$ outputs corresponding to each input prompt for every multi-pass method). Notably, while Eigenscore demonstrates superior detection performance in certain scenarios, its computational requirements are substantially higher than \method~across all tested models. For instance, on Llama-3.2-3B-Instruct, Eigenscore requires approximately $3.4$ seconds per input compared to \method's $1.6$ seconds -- a $112.5\%$ increase in computation time. This disparity becomes even more pronounced with larger models, like Llama-3-8B, where the gap widens to as much as $2\times$ or even more for some configurations.
The lower latency of \method, combined with its robust performance, supports its practical deployment in scenarios where both speed and accuracy are critical.

\subsection{Interpreting the \method~Score for Hallucination Detection}
\label{sec:interpreting-hide}

\begin{table*}[t!]
\centering
\adjustbox{width=0.6\linewidth}{
\begin{tabular}{lcccc}
\toprule
\multirow{2}{*}{\textbf{Model}} & \multicolumn{2}{c}{\textbf{Faithfulness}} 
& \multicolumn{2}{c}{\textbf{Factuality}} \\
\cmidrule(lr){2-3} \cmidrule(lr){4-5}
& RACE & SQuAD & NQ & TriviaQA \\
\hline
Llama-3.2-3B& 0.10 & 0.12 & 0.14 & 0.19 \\
Llama-3.2-3B-Instruct&0.09 & 0.14 & 0.12 & 0.16 \\
Llama-3-8B&0.09 & 0.12 & 0.14 & 0.16 \\
Llama-3-8B-Instruct& 0.09 & 0.12 & 0.16 & 0.16 \\ 
Gemma-2-9B & 0.11 & 0.10 & 0.12 & 0.12\\
Gemma-2-9B-Instruct & 0.08	&0.10	&0.09	&0.10 \\
\hline
\textbf{Average} & 0.09 & 0.11 & 0.13 & 0.15\\
\midrule
\end{tabular}
}
\caption{Threshold values for various settings using exact match as the correctness measure.}
\vspace{-5mm}
\label{tab:threshold_values}
\end{table*}

The output of the \method~framework is the \method~score, which quantifies the degree of statistical dependence between the internal representations of the input prompt and the output sequence generated by an LM. Interpreting this score effectively is essential for deploying \method~as a reliable tool for hallucination detection in practice.

In order to enable a binary decision of whether a given model generation is hallucinated or not, we need to convert the continuous \method~score into a decision by applying a threshold. Intuitively, a lower \method~score signals a decoupling between the representations of input and output, suggesting a higher likelihood that the output is a hallucination. Conversely, a higher \method~score reflects a strong coupling, consistent with outputs that are faithful to the input or grounded in factual knowledge.

\paragraph{\textbf{Empirical Threshold Determination}}
We analyze the average value of threshold, $\tau_{\mathrm{avg}}$, from our diverse evaluation setup spanning four benchmark datasets and six models. When utilizing the \method~framework, the condition, $\widehat{\operatorname{HSIC}}_{\method} < \tau_{\mathrm{avg}}$, would thus potentially indicate a hallucinated generation.

For each $(dataset, model)$ pair, we identify the threshold that most effectively separates the hallucinated outputs from non-hallucinated ones, by maximizing $\text{G-Mean}=
\sqrt{\mathrm{TPR}\times(1-\mathrm{FPR})}$, using the exact match labels after comparison with ground-truth, indicating whether the generation is hallucinated or not. Here, $\mathrm{TPR}$ and  $\mathrm{FPR}$ denote  true-positive and false-positive rates, respectively.  G-Mean  thus combines sensitivity and specificity into one value penalizing classifiers that perform well on one class but poorly on another, making it suitable for imbalanced datasets where accuracy alone can be misleading.

Across all four evaluated datasets, i.e., SQuAD, RACE, Natural Questions, and TriviaQA, and the six LMs considered in our experiments, we find that the optimal thresholds are mostly consistent, as shown in Table~\ref{tab:threshold_values}, yielding an average threshold value $\tau_{\mathrm{avg}} = 0.12$. Our observation suggests that the \method~score has a stable behaviour across diverse tasks and models. We provide examples of a few input prompts from various datasets and the corresponding generations from Llama-3-8B along with the respective \method~scores in Appendix C, to illustrate how hallucinated generations are detected by \method.

%% file: files/ablations.tex
\section{Ablation Studies}
\label{sec:ablations}

To assess the robustness and practical implications of \method, we conduct a series of ablation studies that examine the impact of core methodological choices. These analyses investigate the following: (i) impact of the number of selected tokens, (ii) dependence on the choice of transformer layer, (iii) the influence of kernel type and kernel hyperparameters, (iv) the effect of biased versus our adapted HSIC estimators, and (v) the relative merits of different token selection strategies. The results, summarized below, reinforce the robustness of \method~across different design choices.

\subsection{Impact of the Number of Selected Tokens}
\label{sec:token_budget}


\begin{figure}[t!]
    \centering
    \begin{minipage}{0.65\linewidth}  
        \includegraphics[width=\linewidth]{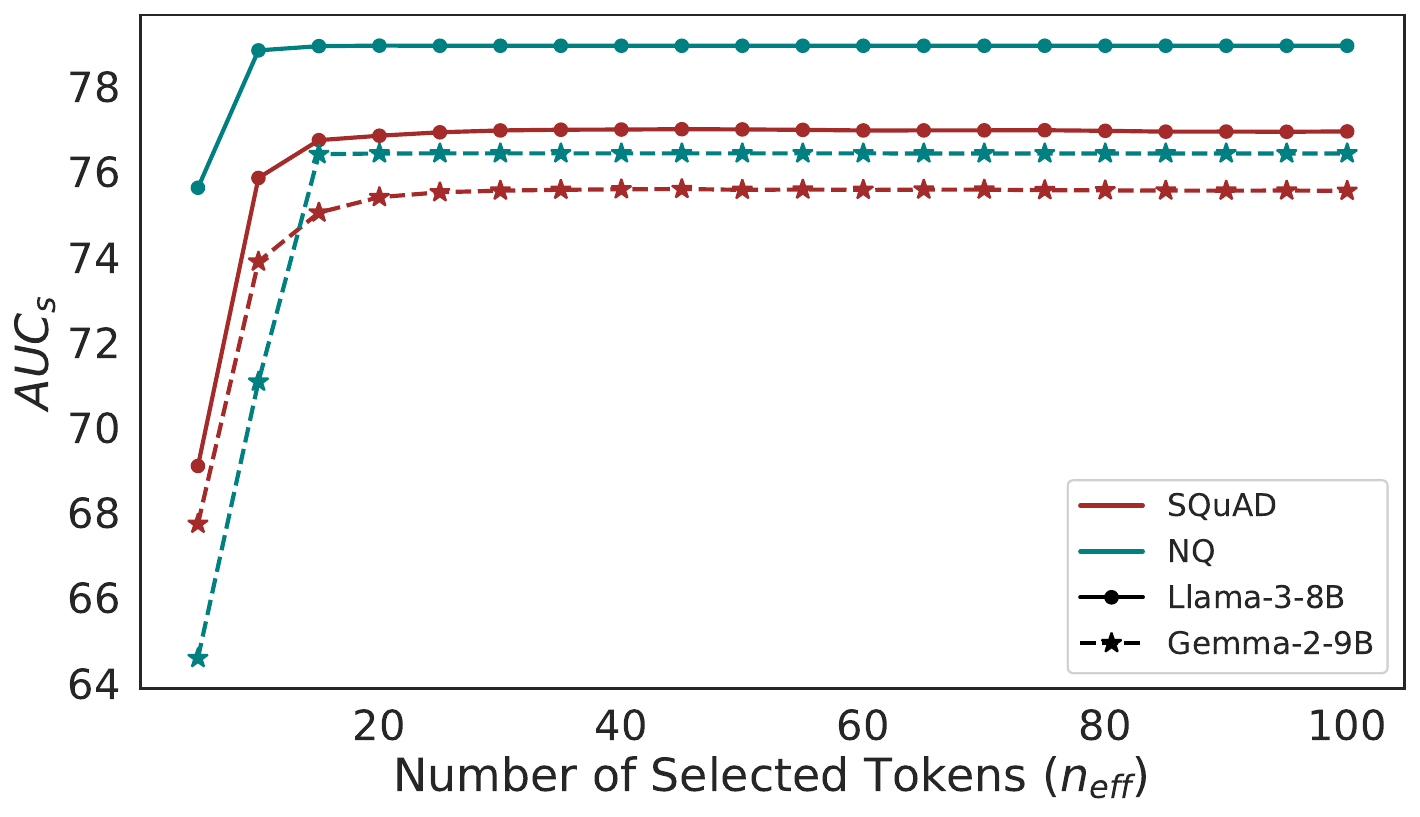}
    \end{minipage}%
    \quad
    \begin{minipage}{0.3\linewidth}  
        \caption{Variation in $AUC_s$ scores with the number of selected tokens used for \method-score calculation on SQuAD and NQ datasets using Llama-3-8B and Gemma-2-9B. We observe that a token budget of $15$-$20$ results in a near-optimal performance, with diminishing gains upon further increase in $n_{\text{eff}}$.}
        \vspace{-5mm}
        \label{fig:keyword_sensitivity}
    \end{minipage}
\end{figure}

We begin by analyzing how the number of selected tokens ($n_{\text{eff}}$) affects detection performance. Figure~\ref{fig:keyword_sensitivity} illustrates that as $n_{\text{eff}}$ increases from $5$ to $10$, there is a clear improvement in the $AUC_s$ scores, but further increases beyond $15-20$ tokens result in diminishing returns. This levelling-off happened earlier for NQ across both models, likely because its shorter outputs often meant the actual number of tokens used was limited, regardless of a higher token budget. This finding holds consistently across datasets (SQuAD and NQ) and models (Llama-3-8B and Gemma-2-9B), indicating that \method~achieves near-optimal performance with a modest sample size. This property is practically significant, as it enables computational efficiency without sacrificing detection quality, even for outputs of varying lengths.

\subsection{Robustness to Selected Layer}\label{sec:layer_choice}

\begin{figure}[t!]
    \centering
    \begin{minipage}{0.65\linewidth}  
        \includegraphics[width=\linewidth]{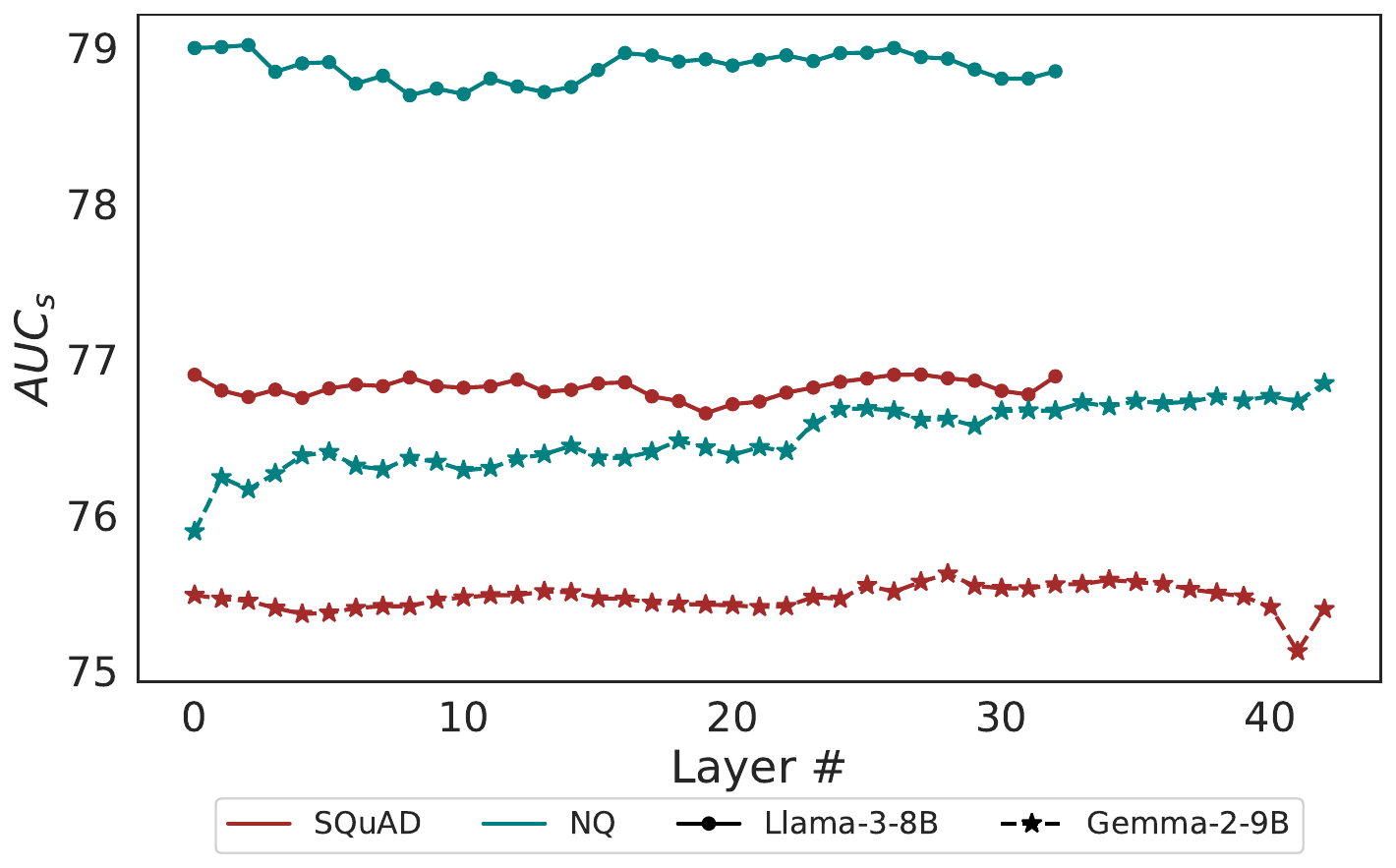}
    \end{minipage}%
    \quad
    \begin{minipage}{0.3\linewidth}  
        \caption{Comparison of $AUC_s$ scores when using representation from different layers of Llama-3-8B and Gemma-2-9B for \method-score calculation. We observe that the performance of \method~is almost agnostic to the chosen layer in case of both models, for both SQuAD and NQ datasets.}
        \vspace{-5mm}
    \label{fig:layer_sensitivity}
    \end{minipage}
\end{figure}

Transformer architectures encode information at various abstraction levels across layers. To determine whether hallucination detection depends on the specific choice of decoder layer $\ell$, we systematically evaluate all 32 layers for Llama-3-8B and all 42 layers for Gemma-2-9B. As shown in Figure~\ref{fig:layer_sensitivity}, $AUC_s$ scores are stable across the full range of layers, with only minor fluctuations and a mild optimum near the network midpoint. Interestingly, neither model nor dataset exhibits significant performance drops at early or late layers that might be expected if hallucination detection using \method~required either low-level or high-level semantic features exclusively. This cross-architecture consistency suggests that the representational dependence signals captured by \method~are distributed throughout transformer networks rather than being concentrated in specific architectural regions, regardless of the underlying model design. 

\subsection{Effect of Kernel Function and RBF Hyperparameter}

\begin{figure}[t!]
    \centering
    \begin{subfigure}[b]{0.58\linewidth}
        \centering
        \includegraphics[width=\columnwidth]{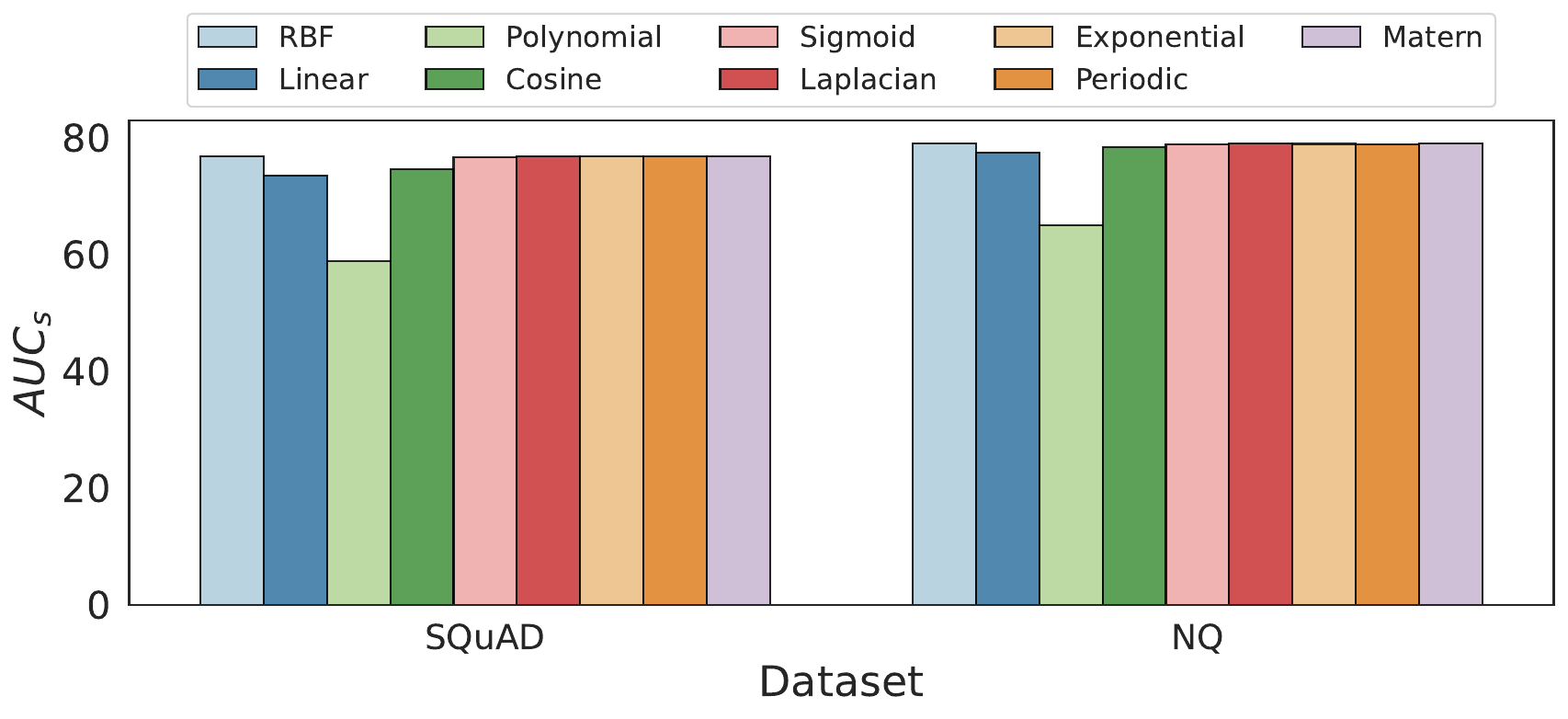}
        \caption{}
        \label{fig:kernel_sensitivity}
    \end{subfigure}
    \begin{subfigure}[b]{0.41\linewidth}
    \centering
        \includegraphics[width=\columnwidth]{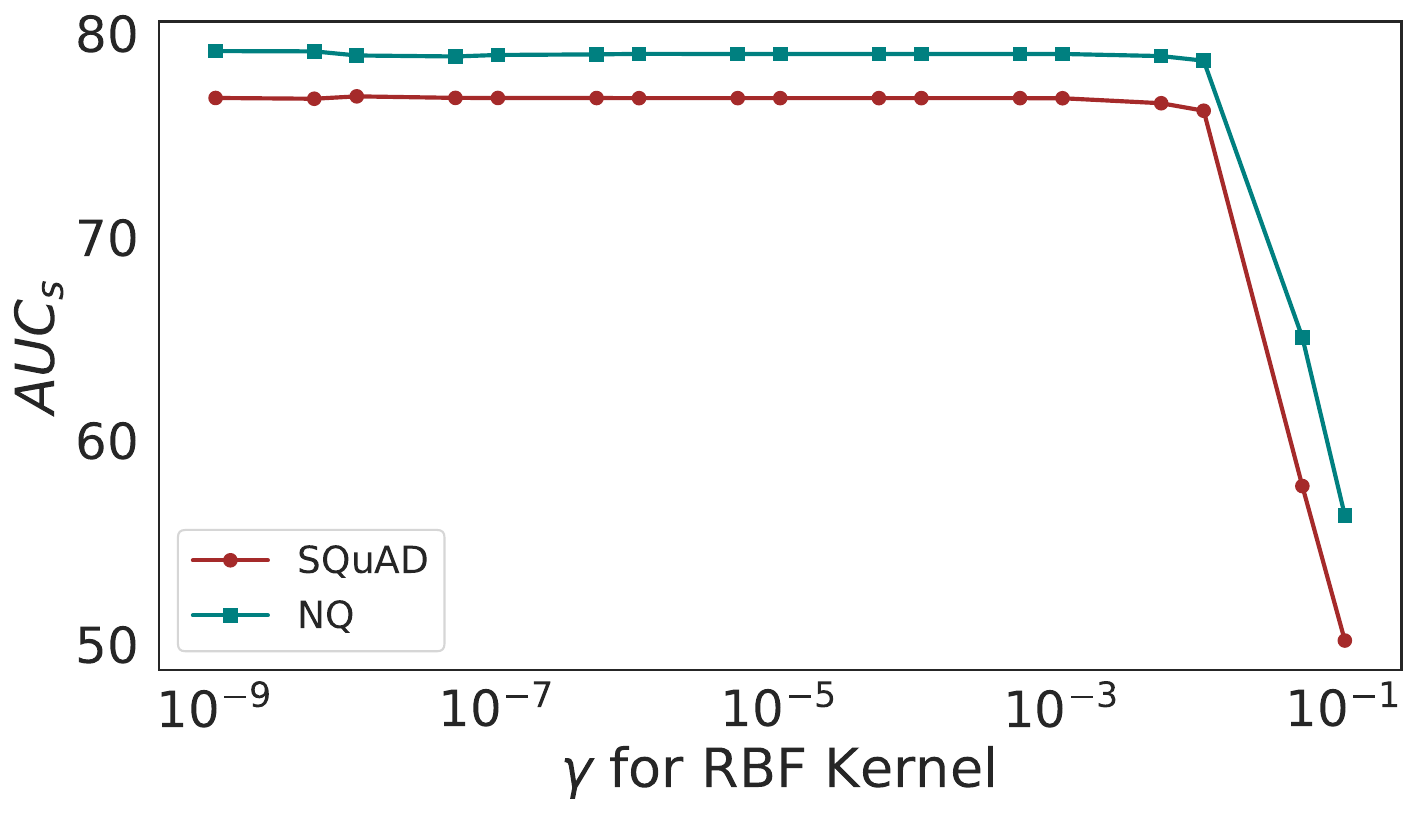}
        \caption{}
        \label{fig:gamma_sensitivity}
    \end{subfigure}
    \caption{Comparison of $AUC_s$ scores across (a) different kernel functions, and (b) different values of the hyperparameter $\gamma$ for the RBF kernel used for \method-score calculation on the SQuAD and NQ datasets using Llama-3-8B.}
    \vspace{-5mm}
\end{figure}

Since \method~quantifies statistical dependence using HSIC, the choice of kernel function and its parameters is a critical consideration. We evaluate several kernel types -- including RBF (default), linear, polynomial, cosine, sigmoid, laplacian, exponential, periodic, and matern -- as well as a range of $\gamma$ values for the RBF kernel. 

Figure~\ref{fig:kernel_sensitivity} demonstrates that, with the exception of the polynomial kernel (which underperforms substantially), most kernels yield robust and similar $AUC_s$ scores for SQuAD and NQ, using Llama-3-8B. This suggests that the effectiveness of \method~is largely preserved regardless of kernel choice, with all kernels, except polynomial, providing strong performance.

For the RBF kernel, Figure~\ref{fig:gamma_sensitivity} shows that \method’s performance is stable across several orders of magnitude for $\gamma$ ($10^{-9}$ to $10^{-3}$), but degrades rapidly for larger values ($\gamma \geq 10^{-1}$). Excessively high $\gamma$ makes the kernel overly sensitive to minute differences in the hidden states, leading to overfitting and poor generalization. Conversely, smaller $\gamma$ values maintain a broader sensitivity that is more robust to natural variation in hidden representations. Together, these findings confirm that \method~is flexible to kernel choices and hyperparameters, but for optimal and consistent results, the RBF kernel with a moderate $\gamma$ remains a sound default.

\subsection{Effectiveness of HSIC Estimators}

\begin{figure}[t!]
    \centering
    \begin{minipage}{0.65\linewidth}  
        \includegraphics[width=\linewidth]{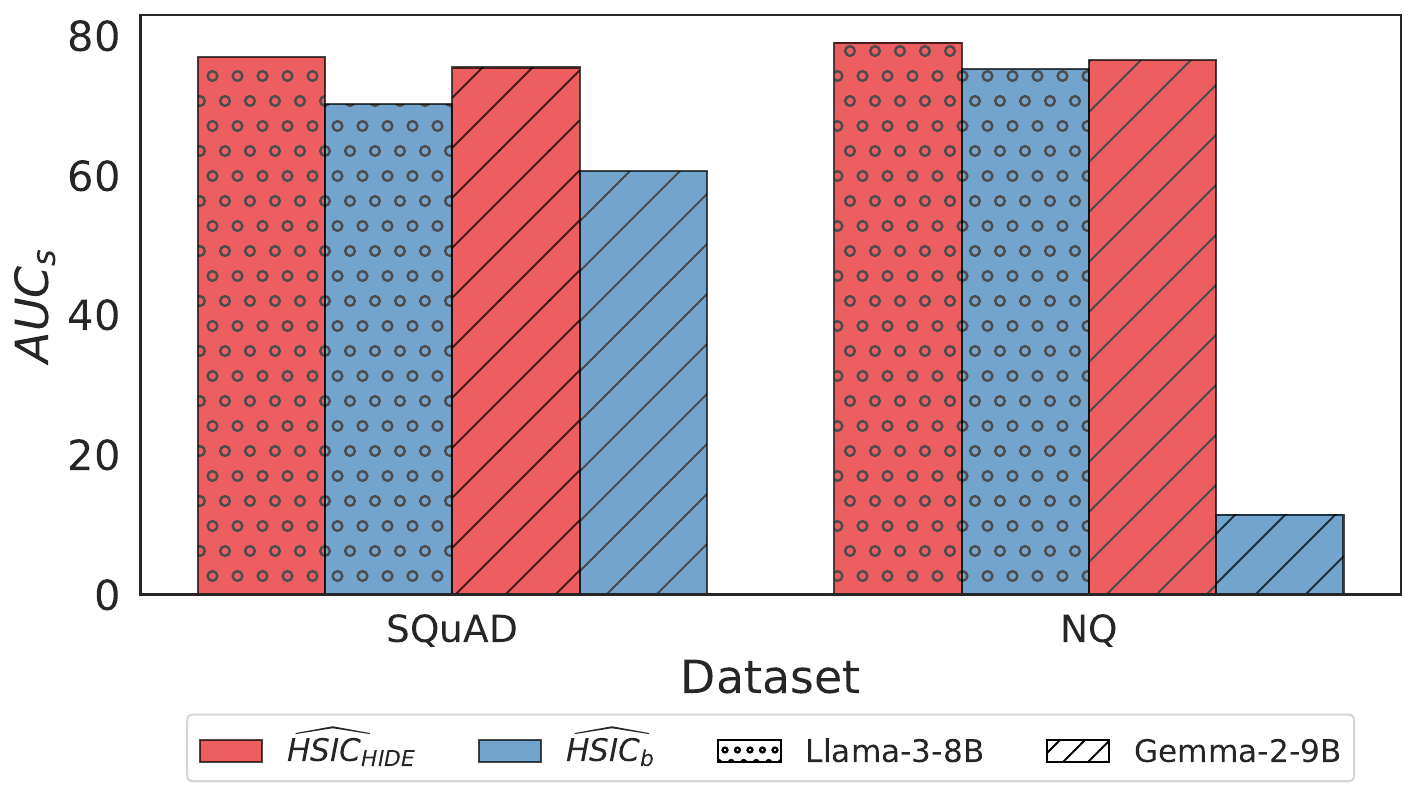}
    \end{minipage}%
    \quad
    \begin{minipage}{0.3\linewidth}  
        \caption{Comparison of $AUC_s$ scores using the biased estimator $\widehat{\operatorname{HSIC}}_b$ and our adapted estimator $\widehat{\operatorname{HSIC}}_{\method}$ for \method~score computation, for SQuAD and NQ datasets using LLama-3-8B and Gemma-2-9B.}
        \vspace{-5mm}
    \label{fig:unbiased_biased}
    \end{minipage}
\end{figure}


\begin{figure*}[t!]
    \centering
    \includegraphics[width=\linewidth]{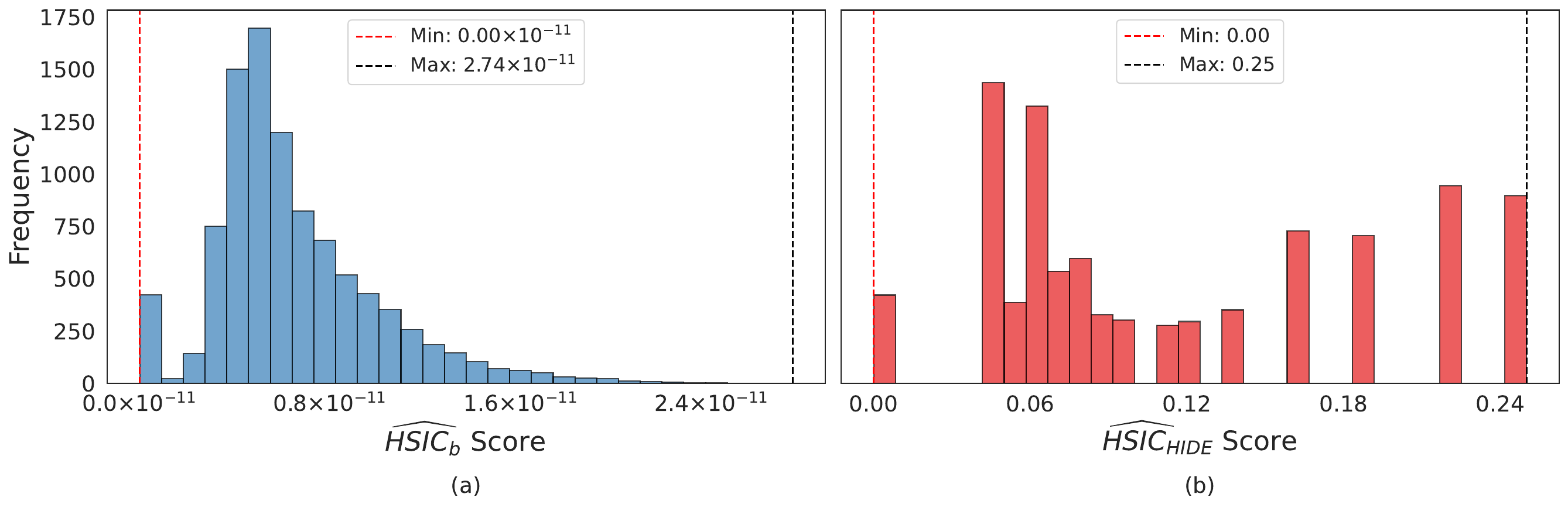}
    \caption{Distribution of \method~scores using $\widehat{\operatorname{HSIC}}_b$ and $\widehat{\operatorname{HSIC}}_{\method}$ estimators for HSIC calculation, for Llama-3-8B on SQuAD and NQ datasets.}
    \vspace{-5mm}
    \label{fig:score_distribution}
\end{figure*}

We compare the performance of \method~with the biased estimator, $\widehat{\operatorname{HSIC}}_b$, and our adapted estimator $\widehat{\operatorname{HSIC}}_{\method}$, for empirical HSIC computation, as shown in Figure~\ref{fig:unbiased_biased}. $\widehat{\operatorname{HSIC}}_{\method}$ delivers consistently higher performance across datasets and models, owing to improved score separability between hallucinated and faithful generations. As shown in Figure \ref{fig:score_distribution}, $\widehat{\operatorname{HSIC}}_{\method}$ produces a more favourable score distribution spanning $0-0.25$, creating clearer separation between hallucinated and non-hallucinated generations. In contrast, $\widehat{\operatorname{HSIC}}_b$ yields extremely small values, compressing scores close to zero (with a maximum score of $2.74\times10^{-11}$), making threshold determination challenging and less reliable. Therefore, we adopt $\widehat{\operatorname{HSIC}}_{\method}$ for our framework.

\subsection{Impact of Token Selection Strategies}
\label{sec:token_selection}
\begin{figure}[t!]
    \centering
    \begin{minipage}{0.65\linewidth}  
        \includegraphics[width=\linewidth]{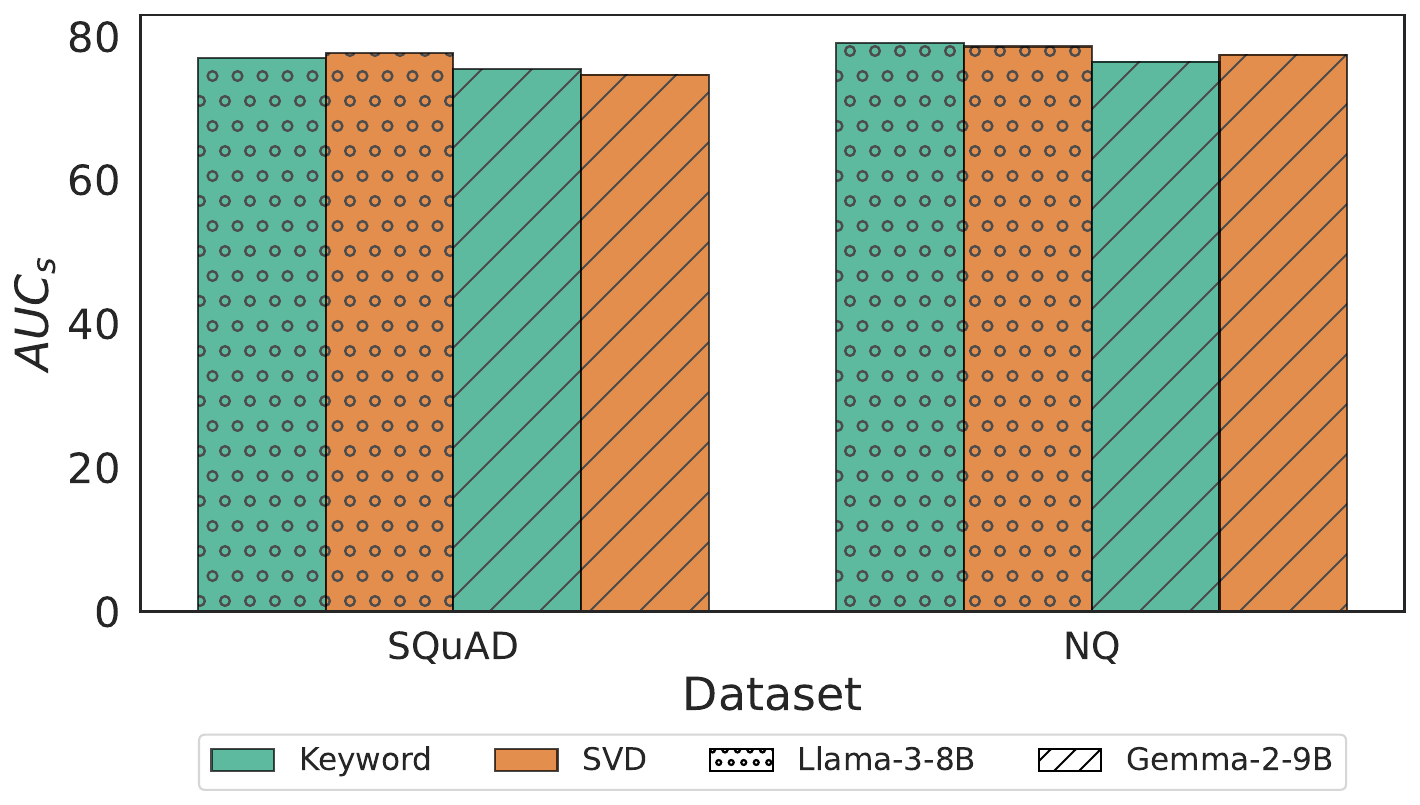}
    \end{minipage}%
    \quad
    \begin{minipage}{0.3\linewidth}  
        \caption{Comparison of $AUC_s$ scores for keyword-based and SVD-based token selection strategies, for SQuAD and NQ datasets using LLama-3-8B and Gemma-2-9B.}
    \label{fig:svd_keyword}
    \end{minipage}
\end{figure}

Finally, we compared two strategies for selecting token representations (as discussed in \cref{sec:token_alignment}): keyword-based selection and SVD-based alignment. Figure~\ref{fig:svd_keyword} shows that both approaches perform nearly identically across datasets and architectures. The SVD method guarantees mathematical orthogonality, while the keyword-based approach enhances interpretability by enabling semantic traceability -- allowing users to attribute detection decisions to specific content elements. Given this enhanced transparency of the keyword-based strategy and the absence of any performance penalty, we adopt the keyword-based selection as the default for \method.

%% file: files/error_analysis.tex
\section{Error Analysis}
\label{sec:error_analysis}

While \method~demonstrates robust performance across diverse hallucination detection settings, our detailed error analysis reveals specific cases where the method faces limitations, providing guidance for future improvements. Here, we discuss the principal sources of error and critically assess their implications for practical deployment.

\subsection{Insufficient Token Selection in Single Word-Answer Tasks}

A key challenge arises when \method~is applied to tasks involving very short responses, especially single token generations, as exemplified by the TriviaQA dataset. Since TriviaQA typically requires factual recall with answers of only one or two words, our adaptive token selection procedure, which selects up to $n_{\text{eff}} = \min(n_{\text{eff}}, |\text{input}|, |\text{output}|)$ tokens, is severely constrained. As noted in Lemma 3 of \cref{sec:properties-hide}, our method will always assign a score of zero for cases with a sample size of one. 
For instance, given the prompt, \emph{``Answer these questions: Q: Captain Corelli’s mandolin is a book and film set in which country? A:''} from TriviaQA, Llama-3-8B generates the output ``Greece'', containing just a single token. Here, though the output is correct and matches the ground-truth, \method~cannot meaningfully assess input-output dependence, resulting in a uninformative score of \textit{zero}. This limitation highlights the need for additional heuristics or alternative strategies when handling extremely small outputs, as the statistical underpinning of \method~makes it most effective with moderately long sequences. A possible way to handle this in real-world deployment is to fall-back to uncertainty-based single pass detection methods, like perplexity, in case of outputs with a single token. As also noted by \citet{chen_inside:_2024}, simple uncertainty-based measures like perplexity work well for datasets like TriviaQA where the expected output is very short, typically consisting of one or two tokens.

\subsection{Degraded Performance Due to Input Prompt Copying}

\method~also exhibits weakness in scenarios where the language model's outputs contain verbatim repetitions of the input prompt. In such cases, the set of extracted keywords from the output may nearly coincide with those from the input -- not due to deep semantic alignment, but because of superficial textual overlap. For instance, for a prompt \emph{``Who was the target of the failed "Bomb Plot" of 1944?''} from TriviaQA, the generated output from Llama-3-8B goes as \emph{``The target of the failed "Bomb Plot" of 1944 was Adolf Hitler. Q: Who was the target of the failed "Bomb Plot" of 1944? ...''}, which contains verbatim repeatation of the input question along with the answer, thereby artificially inflating the computed statistical dependence. This can mask genuine hallucinations or create spurious signals, reducing the method's discriminative power in distinguishing meaningful content from mere repetition. Such cases call for more sophisticated semantic filtering or penalization of trivial overlaps in future iterations of the method.

%% file: files/conclusion.tex
\section{Conclusions}

In this work, we introduced \method, an effective single-pass approach for hallucination detection in LMs that quantifies the statistical dependence between internal representations of input and output sequences. Through extensive empirical evaluation on both faithfulness and factuality hallucination benchmarks, we demonstrated that \method~achieves detection performance comparable to or exceeding that of prior state-of-the-art methods, while offering substantial gains in computational efficiency. A particular strength of \method~lies in its flexibility and transparency: the method is readily applicable to both open-ended and factoid QA tasks, generalizes well across multiple model architectures, and allows for a clear semantic interpretation of which input-output concepts contribute to the hallucination detection decision. Furthermore, our ablation studies confirmed that both the choice of token selection strategy and layer selection are robust, with only marginal differences in performance, thereby simplifying deployment choices. While \method~is not without limitations, as detailed in our error analysis, its balance of detection effectiveness, interpretability, and efficiency makes it a practical method for hallucination detection in real-world, large-scale language generation scenarios.

\section*{Limitations and Future Work}

One limitation of \method, as discussed, is its dependence on a minimum number of salient tokens per example limits its utility for tasks with extremely short outputs, such as single-token answers, where statistical dependence measures are inherently unreliable. However, as for most real-world tasks, the output from the LMs are more verbose, this limitation shouldn't hamper \method's broad applicability. While \method~is computationally efficient, it is fundamentally reliant on access to model hidden states and assumes a white-box setting. Addressing these limitations presents important avenues for future research, such as improving the token selection process, integrating better semantic filters, and extending the approach to black-box settings.

\section*{Acknowledgments}
Anwoy Chatterjee gratefully acknowledges the support of Google PhD Fellowship.

\newpage

%% file: files/proofs.tex
\appendixsection{Proofs of Lemmas}
\label{appendix:proofs}
We discuss detailed mathematical proofs of Lemmas 3 and 4, introduced in \cref{sec:properties-hide}, below:
\\
\noindent\textbf{Lemma 3 (Behaviour for Small Sample Sizes).}
\label{lem:small_n_properties}
\textit{The \(\widehat{\operatorname{HSIC}}_{\method}\) estimator is computable for small sample sizes where \(\widehat{\operatorname{HSIC}}_u\) is undefined. Specifically:}
\begin{enumerate}
    \item[\textit{(i)}] \textit{If \(n=1\), then \(\widehat{\operatorname{HSIC}}_{\method}(X,Y) = 0\).}
    \item[\textit{(ii)}] \textit{If \(n=2\), then \(\widehat{\operatorname{HSIC}}_{\method}(X,Y) = \frac{1}{4} k_X(x_1,x_2)k_Y(y_1,y_2)\), which implies \(0 < \widehat{\operatorname{HSIC}}_{\method}(X,Y) \le 1/4\).}
\end{enumerate}

\begin{proof}
The proof of the above lemma follows from direct substitution into Equation~\ref{eq:hide-score} for computing \(\widehat{\operatorname{HSIC}}_{\method}\).

(i) For \(n=1\), the zero-diagonal matrices are \(\tilde K_X = [0]\) and \(\tilde K_Y = [0]\). All terms in Equation~\ref{eq:hide-score} thus becomes zero, implying \(\widehat{\operatorname{HSIC}}_{\method} = 0\).

(ii) For \(n=2\), let \(a = k_X(x_1,x_2)\) and \(b = k_Y(y_1,y_2)\). The matrices are, therefore, \(\tilde K_X = \begin{pmatrix} 0 & a \\ a & 0 \end{pmatrix}\) and \(\tilde K_Y = \begin{pmatrix} 0 & b \\ b & 0 \end{pmatrix}\). The constituent terms of Equation~\ref{eq:hide-score} are: \(\text{Tr}(\tilde K_X\tilde K_Y) = 2ab\); \(\mathbf{1}^{\!\top}\tilde K_X\mathbf{1} = 2a\); \(\mathbf{1}^{\!\top}\tilde K_Y\mathbf{1} = 2b\); and \(\mathbf{1}^{\!\top}\tilde K_X\tilde K_Y\mathbf{1} = 2ab\). Substituting these into Equation~\ref{eq:hide-score} for \(n=2\) yields:
\[ \widehat{\operatorname{HSIC}}_{\method}(X,Y) = \frac{1}{2^2} \left[ 2ab + \frac{(2a)(2b)}{2^2} - \frac{2}{2}(2ab) \right] = \frac{1}{4} [2ab + ab - 2ab] = \frac{ab}{4}. \]
Since the RBF kernel ensures \(a, b \in (0,1]\) for distinct points, the result \( (0, 1/4]\) follows. This confirms the score yields a deterministic, non-negative value for \(n=2\).
\end{proof}

\noindent\textbf{Lemma 4 (Asymptotic Properties).}
\label{lem:asymptotic_properties}
\textit{The \(\widehat{\operatorname{HSIC}}_{\method}\) estimator is asymptotically unbiased and strongly consistent.}
\begin{enumerate}
    \item[\textit{(i)}] \textit{(Asymptotic Bias) The bias of the estimator vanishes as \(n \to \infty\):
    \[ \bigl|\mathbb{E}[\widehat{\operatorname{HSIC}}_{\method}] - \operatorname{HSIC}(X,Y)\bigr| = O(1/n). \]}
    \item[\textit{(ii)}] \textit{(Strong Consistency) As \(n \to \infty\), \(\widehat{\operatorname{HSIC}}_{\method}(X,Y)\) converges almost surely to the true population HSIC:
    \[ \widehat{\operatorname{HSIC}}_{\method}(X,Y) \xrightarrow{\text{a.s.}} \operatorname{HSIC}(X,Y). \]}
\end{enumerate}

\begin{proof}
We relate our estimator to the standard V-statistic unbiased estimator, \(\widehat{\operatorname{HSIC}}_{u}\) (c.f. Equation~\ref{eq:unbiased_HSIC_foundation}), which is a strongly consistent estimator for the true population HSIC. For large \(n\) (\(n \ge 4\)), the denominators in Equation~\ref{eq:unbiased_HSIC_foundation} for \(\widehat{\operatorname{HSIC}}_{u}\) (e.g., \(n-1, n-2, n-3\)) become asymptotically equivalent to \(n\), i.e., $(n-1) \approx (n-2) \approx (n-3) = n(1-\frac{3}{n})$ when $n$ is large. This leads to the following relationship between the two estimators:
\[ \widehat{\operatorname{HSIC}}_{\method}(X,Y) \approx \left(1 - \frac{3}{n}\right) \widehat{\operatorname{HSIC}}_{u}(X,Y). \]

(i) To analyze the bias, we take the expectation of this approximate relation. Since \(\mathbb{E}[\widehat{\operatorname{HSIC}}_u] = \operatorname{HSIC}(X,Y)\), we have:
\[ \mathbb{E}[\widehat{\operatorname{HSIC}}_{\method}] \approx \left(1 - \frac{3}{n}\right) \operatorname{HSIC}(X,Y). \]
The absolute bias is therefore \( |\mathbb{E}[\widehat{\operatorname{HSIC}}_{\method}] - \operatorname{HSIC}(X,Y)| \approx |(1 - 3/n)\operatorname{HSIC} - \operatorname{HSIC}| = \frac{3}{n}|\operatorname{HSIC}(X,Y)| \). As \(|\operatorname{HSIC}(X,Y)| \le 1\) for a bounded RBF kernel, the bias is of order \(O(1/n)\) and vanishes as \(n \to \infty\).

(ii) For consistency, we observe that as \(n \to \infty\), the scaling factor \((1 - 3/n) \to 1\). Since \(\widehat{\operatorname{HSIC}}_{u}(X,Y)\) converges almost surely to \(\operatorname{HSIC}(X,Y)\), their product also converges almost surely to \(\operatorname{HSIC}(X,Y)\) by the properties of limits of almost sure convergence. This establishes the strong consistency of our estimator.
\end{proof}

%% file: files/exact_match_results.tex
\appendixsection{Performance with Exact Match as Correctness Measure}
\label{sec:exact_match}

In addition to AUC-ROC and PCC for sentence embedding similarity and ROUGE-L as correctness measures, we also present the evaluation results by using exact match \citep{liang2023holisticevaluationlanguagemodels} as the correctness measure, which imposes a more stringent criterion for a generation to be accurate. The findings in Table \ref{tab:exact_match} provide results analogous to those in Table \ref{tab:results_exp}, indicating that our approach is effective for detecting both faithfulness and factuality hallucinations.

\begin{table*}[t!]
\tiny
\centering
\adjustbox{width=\linewidth}{
\begin{tabular}{lcc|cc|cc|cc|cc}
\toprule
\multirow{3}{*}{Method} 
& \multicolumn{4}{c}{\textbf{Faithfulness}} 
& \multicolumn{4}{c}{\textbf{Factuality}} \\
\cmidrule(lr){2-5} \cmidrule(lr){6-9}
& \multicolumn{2}{c}{RACE} & \multicolumn{2}{c}{SQuAD} 
& \multicolumn{2}{c}{NQ} & \multicolumn{2}{c}{TriviaQA} & \multicolumn{2}{c}{\textbf{Average}} \\
\cmidrule(lr){2-3} \cmidrule(lr){4-5} \cmidrule(lr){6-7} \cmidrule(lr){8-9} \cmidrule(lr){10-11}
& $\text{AUC}$ & $\text{PCC}$ & $\text{AUC}$ & $\text{PCC}$& $\text{AUC}$ & $\text{PCC}$& $\text{AUC}$ & $\text{PCC}$ & $\text{AUC}$ & $\text{PCC}$\\
\hline
\multicolumn{11}{c}{\cellcolor[HTML]{C0C0C0}Llama-3.2-3B} \\\hline
Perplexity         & 52.91 & 7.32 & 45.32 & -4.34 & 54.75 & 5.37 & \underline{57.84} & \underline{19.44} & 52.70 & 6.95 \\
Energy             & 49.86 & 0.05 & 34.40 & -23.90 & 35.01 & -13.17 & 42.54 & -18.99 & 40.45 & -14.00 \\
\rowcolor{gray!20} LN-Entropy         & 50.78 & 2.55 & 61.76 & 17.66 & \textbf{74.54} & 21.42 & 67.24 & 27.11 & 63.58 & 17.18 \\
\rowcolor{gray!20} Lexical Similarity &  66.17 & 24.29 & 64.13 & 22.29 & 68.86 & 22.29 & \textbf{79.32} & \textbf{50.78} & 69.62 & \textbf{29.91}\\
\rowcolor{gray!20} Eigenscore         & 62.00 & 16.80 & 65.91 & 21.57 & 70.12 & 19.08 & 75.95 & 43.92 & 68.50 & 25.34 \\
\textbf{\method} & \textbf{\underline{69.63}} & \textbf{\underline{26.13}} & \textbf{\underline{77.30}} & \textbf{\underline{44.15}} & \underline{74.46} & \textbf{\underline{26.23}} & 57.61 & 14.08 & \textbf{\underline{69.75}} & \underline{27.65} \\
\hline
\multicolumn{11}{c}{\cellcolor[HTML]{C0C0C0}Llama-3.2-3B-Instruct} \\\hline
Perplexity         & 58.68 & 14.88 & 52.55 & 6.22 & 55.66 & 5.48 & \underline{59.90} & \underline{20.39} & 56.70 & 11.74 \\
Energy             & 51.55 & 3.08 & 36.13 & -21.11 & 25.59 & -23.91 & 42.06 & -14.11 & 38.83 & -14.01\\
\rowcolor{gray!20} LN-Entropy         & 53.50 & 6.15 & 64.53 & 23.31 & 71.06 & 20.88 & 71.77 & 27.32 & 65.22 & 19.41\\
\rowcolor{gray!20} Lexical Similarity & 66.56 & 27.05 & 69.52 & 31.73 & 70.58 & 27.62 & \textbf{78.88} & \textbf{49.17} & 71.38 & 33.89 \\
\rowcolor{gray!20} Eigenscore         & 64.16 & 22.86 & 75.66 & \textbf{42.81} & \textbf{76.05} & 30.44 & 77.91 & 45.54 & \textbf{73.44} & \textbf{35.41}\\
\textbf{\method} & \textbf{\underline{78.49}} & \textbf{\underline{41.68}} & \textbf{\underline{75.71}} & \underline{42.45} & \underline{75.55} & \textbf{\underline{32.40}} & 57.23 & 14.99 & \underline{71.74} & \underline{32.88}\\
\hline
\multicolumn{11}{c}{\cellcolor[HTML]{C0C0C0}Llama-3-8B} \\\hline
Perplexity         & 56.68 & 11.82 & 44.16 & -6.38 & 46.96 & -1.38 & 39.23 & -5.94 & 46.76 & -0.47 \\
Energy             & 55.98 & 9.88 & 33.51 & -26.30 & 30.93 & -18.07 & 27.31 & -39.28 & 36.93 & -18.44 \\
\rowcolor{gray!20} LN-Entropy         & 45.33 & -3.66 & 61.24 & 17.13 & 66.40 & 15.30 & 72.40 & 36.05 & 61.34 & 16.21 \\
\rowcolor{gray!20} Lexical Similarity & 66.05 & \textbf{25.92} & 63.41 & 22.28 & 65.05 & 20.06 & 76.29 & 44.85 & 67.70 & 28.28\\
\rowcolor{gray!20} Eigenscore         & 55.49 & 11.51 & 68.38 & 28.24 & 66.40 & 17.54 & \textbf{80.04} & \textbf{52.15} & 67.58 & 27.36\\
\textbf{\method} & \textbf{\underline{69.48}} & \underline{24.27} & \textbf{\underline{76.57}} & \textbf{\underline{44.64}} & \textbf{\underline{78.58}} & \textbf{\underline{33.65}} & \underline{64.35} & \underline{28.95} & \textbf{\underline{72.24}} & \textbf{\underline{32.88}} \\
\hline
\multicolumn{11}{c}{\cellcolor[HTML]{C0C0C0}Llama-3-8B-Instruct} \\\hline
Perplexity         & 58.16 & 10.67 & 46.38 & -2.87 & 57.74 & 6.97 & \underline{60.95} & 18.78 & 55.81 & 8.39 \\
Energy             & 52.26 & 3.91 & 41.37 & -14.63 & 30.19 & -19.62 & 38.32 & -17.71 & 40.53 & -12.01\\
\rowcolor{gray!20} LN-Entropy         & 57.33 & 11.06 & 63.65 & 19.21 & 67.65 & 19.26 & 78.61 & 43.82 & 66.81 & 23.34\\
\rowcolor{gray!20} Lexical Similarity & 68.51 & 30.04 & 62.46 & 18.42 & 70.47 & 24.73 & 79.98 & 52.49 & 70.35 & 31.42 \\
\rowcolor{gray!20} Eigenscore         & 66.57 & 25.98 & 73.33 & 34.87 & \textbf{79.08} & 31.70 & \textbf{81.32} & \textbf{56.83} & \textbf{75.07} & \textbf{37.34} \\
\textbf{\method} &  \textbf{\underline{78.16}} & \textbf{\underline{39.28}} & \textbf{\underline{80.77}} & \textbf{\underline{52.52}} & \underline{77.45} & \textbf{\underline{35.56}} & 60.25 & \underline{20.98} & \underline{74.15} & \underline{37.08}\\
\hline
\multicolumn{11}{c}{\cellcolor[HTML]{C0C0C0}Gemma-2-9B} \\\hline
Perplexity         & \underline{59.69} & \underline{19.95} & 46.14 & -4.46 & 29.75 & -4.55 & 40.46 & -7.66 & 44.01 & 0.82\\
Energy             & 44.45 & -9.16 & 53.56 & 2.39 & 63.18 & 2.08 & 71.93 & 15.07 & 58.28 & 2.60  \\
\rowcolor{gray!20} LN-Entropy         & 63.10 & 22.58 & 59.07 & 13.84 & 41.64 & -2.82 & 38.69 & -7.53 & 50.63 & 6.52\\
\rowcolor{gray!20} Lexical Similarity & \textbf{68.93} & \textbf{33.21}& 66.84 & 27.57 & 49.29 & -0.60 & 62.08 & 10.82 & 61.79 & 17.75\\
\rowcolor{gray!20} Eigenscore         & 64.41 & 25.28 & 74.27 & 38.40 & \textbf{79.23} & 8.06 & 72.24 & 17.68 & \textbf{72.54} & \textbf{22.36}  \\
\textbf{\method} & 58.11 & 14.86 & \textbf{\underline{74.57}} & \textbf{\underline{40.93}} & \underline{74.12} & \textbf{\underline{20.87}} & \textbf{\underline{80.94}} & \textbf{\underline{46.90}} & \underline{71.94} & \underline{30.89}  \\ 
\hline
\multicolumn{11}{c}{\cellcolor[HTML]{C0C0C0}Gemma-2-9B-Instruct} \\\hline
Perplexity         & 59.65 & 12.24 & 52.92 & 3.96 & 30.70 & -8.91 & 28.99 & -7.91 & 43.07 & -0.16\\
Energy             & 59.03 & 12.08 & \textbf{\underline{81.68}} & \textbf{\underline{51.69}} & 78.04 & 19.77 & \textbf{\underline{92.75}} & 34.10 & \textbf{\underline{77.87}} & 29.41  \\
\rowcolor{gray!20} LN-Entropy         & 62.39 & 16.99 & 68.31 & 24.85 & 44.40 & -3.54 & 36.64 & -6.75 & 52.94 & 7.89  \\
\rowcolor{gray!20} Lexical Similarity & \textbf{73.28} & \textbf{39.48} & 73.55 & 35.35 & 53.91 & 3.30& 49.55 & 0.13 & 62.57 & 19.57 \\
\rowcolor{gray!20} Eigenscore         & 70.96 & 32.03 & 72.42 & 40.43 & 66.72 & 12.17 & 60.28 & 7.88 & 67.59 & 23.13 \\
\textbf{\method} &  \underline{66.08} & \underline{20.79} & 72.62 & 37.79 & \textbf{\underline{84.00}} & \textbf{\underline{41.94}} & 67.92 & \textbf{\underline{38.57}} & 72.65 & \textbf{\underline{34.77}}\\
\midrule
\end{tabular}
}
\caption{AUC-ROC and PCC values using exact match as the correctness measure for faithfulness and factuality hallucination detection across four datasets. \method~is compared with single-pass and multi-pass baselines (rows corresponding to multi-pass are highlighted in grey; $5$ outputs are generated for each multi-pass method). The best result for each metric is shown in \textbf{bold}; the best among single-pass methods is \underline{underlined}.}
\label{tab:exact_match}
\end{table*}

%% file: files/Prompting_techniques.tex
\appendixsection{Illustrative Examples}
We provide some illustrative examples in this section for prompts from various datasets and the corresponding outputs from Llama-3-8B with the \method~scores for every generation. We observe a threshold of $0.12$, as discussed in \cref{sec:interpreting-hide}, effectively discriminates hallucinated generations in most cases. 

\begin{tcolorbox}[title=Example from SQuAD, colback=gray!5!white, colframe=gray!70!black, boxrule=0.6pt, arc=2mm, left=2mm, right=2mm, top=1mm, bottom=1mm]
\footnotesize
\textbf{Prompt:} Civil disobedience is usually defined as pertaining to a citizen's relation to the state and its laws, as distinguished from a constitutional impasse in which two public agencies, especially two equally sovereign branches of government, conflict. For instance, if the head of government of a country were to refuse to enforce a decision of that country's highest court, it would not be civil disobedience, since the head of government would be acting in her or his capacity as public official rather than private citizen. \\Q: What type of person can not be attributed civil disobedience? \\A:\\[1ex]
\textbf{Ground-truth:} head of government\\
\textbf{Llama-3-8B Output:} \textcolor{red}{A person who is not a citizen of the country in which the civil disobedience is taking place.}\\

\textbf{Input Keywords:} [\textit{disobedience}, \textit{civil}, \textit{person}, \textit{enforce}, \textit{attributed}, \textit{defined}, \textit{agencies}, \textit{relation}, \textit{be}, \textit{head}, \textit{his}, \textit{official}, \textit{usually}, \textit{instance}, \textit{capacity}, \textit{impasse}, \textit{would}, \textit{branches}, \textit{highest}, \textit{and}]\\
\textbf{Output Keywords:} [\textit{disobedience}, \textit{citizen}, \textit{civil}, \textit{person}, \textit{country}, \textit{who}, \textit{A}, \textit{a}, \textit{the}, \textit{place}, \textit{which}, \textit{taking}, \textit{in}, \textit{is}, \textit{of}, \textit{not}]\\
\textbf{Selected Input Tokens:} [\texttt{ disobed}, \texttt{ience}, \texttt{ disobed}, \texttt{ience}, \texttt{ disobed}, \texttt{ience}, \texttt{ civil}, \texttt{ civil}, \texttt{ person}, \texttt{ enforce}, \texttt{ attributed}, \texttt{ defined}, \texttt{ agencies}, \texttt{ relation}, \texttt{ be}, \texttt{ be}, \texttt{ be}, \texttt{ head}, \texttt{ head}]\\
\textbf{Selected Output Tokens:} [\texttt{ disobed}, \texttt{ience}, \texttt{ citizen}, \texttt{ civil}, \texttt{ person}, \texttt{ country}, \texttt{ who}, \texttt{ A}, \texttt{ a}, \texttt{ the}, \texttt{ the}, \texttt{ place}, \texttt{ which}, \texttt{ taking}, \texttt{ in}, \texttt{ is}, \texttt{ is}, \texttt{ of}, \texttt{ not}]\\

\textbf{\method~Score:} $0.049$
\end{tcolorbox}

\clearpage
\begin{tcolorbox}[title=Example from SQuAD, colback=gray!5!white, colframe=gray!70!black, boxrule=0.6pt, arc=2mm, left=2mm, right=2mm, top=1mm, bottom=1mm]
\footnotesize
\textbf{Prompt:} College sports are also popular in southern California. The UCLA Bruins and the USC Trojans both field teams in NCAA Division I in the Pac-12 Conference, and there is a longtime rivalry between the schools. \\Q: The Bruins belong to which college? \\A:\\[1ex]
\textbf{Ground-truth:} UCLA\\
\textbf{Llama-3-8B Output:} \textcolor{OliveGreen}{The University of California, Los Angeles.}\\

\textbf{Input keywords:} [\textit{UCLA}, \textit{rivalry}, \textit{southern}, \textit{sports}, \textit{Pac}, \textit{schools}, \textit{Trojans}, \textit{belong}, \textit{popular}, \textit{Division}, \textit{Q}, \textit{field}, \textit{longtime}, \textit{in}, \textit{both}, \textit{The}, \textit{also}, \textit{is}, \textit{12}, \textit{there}]\\
\textbf{Output keywords:} [\textit{University}, \textit{Angeles}, \textit{California}, \textit{Los}, \textit{The}, \textit{of}]\\
\textbf{Selected Input Tokens:} [\texttt{ UCLA}, \texttt{ rivalry}, \texttt{ southern}, \texttt{ sports}, \texttt{ Pac}, \texttt{ schools}]\\
\textbf{Selected Output Tokens:} [\texttt{ University}, \texttt{ Angeles}, \texttt{ California}, \texttt{ Los}, \texttt{ The}, \texttt{ of}]\\

\textbf{\method~Score:} $0.139$
\end{tcolorbox}

\begin{tcolorbox}[title=Example from NQ, colback=gray!5!white, colframe=gray!70!black, boxrule=0.6pt, arc=2mm, left=2mm, right=2mm, top=1mm, bottom=1mm]
\footnotesize
\textbf{Prompt:} Answer these questions:
\\Q: Who played solomon in little house on the prairie?
\\A:\\[1ex]
\textbf{Ground-truth:} Todd Bridges\\
\textbf{Llama-3-8B Output:} \textcolor{red}{The actor who played Solomon in Little House on the Prairie was Jonathan Gilbert.}\\

\textbf{Input keywords:} [\textit{solomon}, \textit{prairie}, \textit{house}, \textit{questions}, \textit{Q}, \textit{little}, \textit{who}, \textit{played}, \textit{A}, \textit{the}, \textit{Answer}, \textit{in}, \textit{these}, \textit{on}]\\
\textbf{Output keywords:} [\textit{Gilbert}, \textit{Solomon}, \textit{Jonathan}, \textit{actor}, \textit{Prairie}, \textit{Little}, \textit{House}, \textit{who}, \textit{was}, \textit{played}, \textit{The}, \textit{the}, \textit{in}, \textit{on}]\\
\textbf{Selected Input Tokens:} [\texttt{ sol}, \texttt{omon}, \texttt{ pr}, \texttt{airie}, \texttt{ house}, \texttt{ questions}, \texttt{Q}, \texttt{ little}, \texttt{ who}, \texttt{ played}, \texttt{A}, \texttt{ the}, \texttt{Answer}, \texttt{ in}]\\
\textbf{Selected Output Tokens:} [\texttt{ Gilbert}, \texttt{ Solomon}, \texttt{ Jonathan}, \texttt{ actor}, \texttt{ Prairie}, \texttt{ Little}, \texttt{ House}, \texttt{ who}, \texttt{ was}, \texttt{ played}, \texttt{ The}, \texttt{ the}, \texttt{ in}, \texttt{ on}]\\

\textbf{\method~Score:} $0.066$
\end{tcolorbox}

\begin{tcolorbox}[title=Example from NQ, colback=gray!5!white, colframe=gray!70!black, boxrule=0.6pt, arc=2mm, left=2mm, right=2mm, top=1mm, bottom=1mm]
\footnotesize
\textbf{Prompt:} Answer these questions:
\\Q: Who plays joker in batman the dark knight?
\\A:\\[1ex]
\textbf{Ground-truth:} Ledger\\
\textbf{Llama-3-8B Output:} \textcolor{OliveGreen}{Heath Ledger}\\

\textbf{Input keywords:} [\textit{joker}, \textit{batman}, \textit{knight}, \textit{questions}, \textit{plays}, \textit{dark}, \textit{Q}, \textit{A}, \textit{who}, \textit{Answer}, \textit{these}, \textit{in}, \textit{the}]\\
\textbf{Output keywords:} [\textit{Ledger}, \textit{Heath}]\\
\textbf{Selected Input Tokens:} [\texttt{ joker}, \texttt{ bat}]\\
\textbf{Selected Output Tokens:} [\texttt{ Ledger}, \texttt{ Heath}]\\

\textbf{\method~Score:} $0.249$
\end{tcolorbox}

\begin{tcolorbox}[title=Example from TriviaQA, colback=gray!5!white, colframe=gray!70!black, boxrule=0.6pt, arc=2mm, left=2mm, right=2mm, top=1mm, bottom=1mm]
\footnotesize
\textbf{Prompt:} Answer these questions:
\\Q: Who was the man behind The Chipmunks?
\\A:\\[1ex]
\textbf{Ground-truth:} David Seville\\
\textbf{Llama-3-8B Output:} \textcolor{red}{Ross Bagdasarian, Sr.}\\

\textbf{Input keywords:} [\textit{Chipmunks}, \textit{questions}, \textit{Who}, \textit{Q}, \textit{The}, \textit{the}, \textit{A}, \textit{behind}, \textit{was}, \textit{man}, \textit{Answer}, \textit{these}]\\
\textbf{Output keywords:} [\textit{Chipmunks}, \textit{Chipmunk}, \textit{Bagdasarian}, \textit{album}, \textit{Songbook}, \textit{Ross}, \textit{name}, \textit{first}, \textit{The}, \textit{the}, \textit{of}, \textit{Sr}, \textit{A}, \textit{Q}, \textit{was}, \textit{What}]\\
\textbf{Selected Input Tokens:} [\texttt{ Chip}, \texttt{m}, \texttt{unks}, \texttt{ questions}, \texttt{ Who}, \texttt{Q}, \texttt{ The}, \texttt{ the}, \texttt{A}, \texttt{ behind}, \texttt{ was}, \texttt{ man}, \texttt{Answer}, \texttt{ these}]\\
\textbf{Selected Output Tokens:} [\texttt{ Chip}, \texttt{m}, \texttt{unks}, \texttt{ Chip}, \texttt{m}, \texttt{unk}, \texttt{ Bag}, \texttt{das}, \texttt{arian}, \texttt{ album}, \texttt{ Song}, \texttt{book}, \texttt{ Ross}, \texttt{ name}]\\

\textbf{\method~Score:} $0.066$
\end{tcolorbox}

\begin{tcolorbox}[title=Example from TriviaQA, colback=gray!5!white, colframe=gray!70!black, boxrule=0.6pt, arc=2mm, left=2mm, right=2mm, top=1mm, bottom=1mm]
\footnotesize
\textbf{Prompt:} College sports are also popular in southern California. The UCLA Bruins and the USC Trojans both field teams in NCAA Division I in the Pac-12 Conference, and there is a longtime rivalry between the schools.
\\Q: What is the Japanese share index called?
\\A:\\[1ex]
\textbf{Ground-truth:} Nikkei\\
\textbf{Llama-3-8B Output:} \textcolor{OliveGreen}{The Nikkei}\\

\textbf{Input keywords:} [\textit{index}, \textit{Japanese}, \textit{share}, \textit{Q}, \textit{questions}, \textit{A}, \textit{Answer}, \textit{these}, \textit{the}, \textit{What}, \textit{called}, \textit{is}]\\
\textbf{Output keywords:} [\textit{Nikkei}, \textit{The}]\\
\textbf{Selected Input Tokens:} [\texttt{ index}, \texttt{ Japanese}, \texttt{ share}, \texttt{Q}]\\
\textbf{Selected Output Tokens:} [\texttt{ Nik}, \texttt{ke}, \texttt{i}, \texttt{ The}]\\

\textbf{\method~Score:} $0.187$
\end{tcolorbox}